%% file: draft.tex
\documentclass[10pt,journal,compsoc]{IEEEtran}

\usepackage[table]{xcolor}
\usepackage{subcaption}
\usepackage{forloop}
\usepackage{etoolbox}

\usepackage[nocompress]{cite}
\usepackage[pdftex]{graphicx}

\usepackage{amsmath}
\usepackage{amssymb}
\usepackage{eucal}

\usepackage{array}
\usepackage{url}

\usepackage[margin=10pt,font=small,labelfont=bf,labelsep=endash]{caption}

\input{macro.tex}

\usepackage{hyperref}

\begin{document}

\title{Exploring Context with Deep Structured models for Semantic Segmentation}

\author{Guosheng Lin, Chunhua Shen, Anton {van den Hengel}, Ian Reid
\IEEEcompsocitemizethanks{
\IEEEcompsocthanksitem
G. Lin is with School of Computer Science and Engineering, Nanyang Technological University, Singapore.
This work was done when G. Lin was with The University of Adelaide and Australian Centre for Robotic Vision. \protect E-mail: guosheng.lin@gmail.com 

C. Shen, A. van den Hengel and I. Reid are with School of Computer Science, The University of Adelaide,
Australia; and Australian Centre for Robotic Vision. 
\protect E-mail: \{chunhua.shen, anton.vandenhengel, ian.reid\}@adelaide.edu.au 

\IEEEcompsocthanksitem
C. Shen is the corresponding author.
}
\thanks{}
}

\markboth{Appearing in IEEE Transactions on Pattern Analysis and Machine Intelligence, April 2017.}{}

\IEEEtitleabstractindextext{%

\begin{abstract}

State-of-the-art semantic image segmentation methods are mostly based on training deep convolutional neural networks (CNNs). In this work, we proffer to improve semantic segmentation with the use of contextual information. In particular, we explore patch-patch context and patch-background context in deep CNNs. We formulate deep structured models by combining CNNs and Conditional Random Fields (CRFs) for learning the patch-patch context between image regions. Specifically, we formulate CNN-based pairwise potential functions to capture semantic correlations between neighboring patches. Efficient piecewise training of the proposed deep structured model is then applied in order to avoid repeated expensive CRF inference during the course of back propagation.For capturing the patch-background context, we show that a network design with traditional multi-scale image inputs and sliding pyramid pooling is very effective for improving performance. We perform  comprehensive  evaluation of the proposed method. We achieve new state-of-the-art performance on a number of challenging semantic segmentation datasets.

\end{abstract}


}

\maketitle

\IEEEdisplaynontitleabstractindextext

\IEEEpeerreviewmaketitle

\tableofcontents
\vspace{2cm}
\clearpage
\vspace{2cm}

\input{intro.tex}

\input{method.tex}

\input{exp.tex}

\section{Conclusions}

We have proposed a method which combines CNNs and CRFs to exploit {\em complex}
contextual information for semantic image segmentation.
Basically, we formulate CNN based pairwise potentials for modeling
semantic relations between image regions.
We have performed comprehensive experiments on $8$ challenging segmentation datasets and
we achieve state-of-the-art performance on all evaluated dataset including the PASCAL VOC 2012 dataset.
The proposed method is potentially widely applicable to other tasks.

\section*{Acknowledgments}
This research was supported  by  Australian Research Council
through the ARC Centre for Robotic Vision (CE140100016).
C. Shen's participation was in part supported by  an ARC Future Fellowship (FT120100969).
I. Reid's participation was in part supported by an ARC Laureate Fellowship (FL130100102).

{
\bibliographystyle{IEEEtran}
\bibliography{CSRef}
}

\end{document}

%% file: macro.tex
\newcommand{\cN}{\mathcal{N}}

\newcommand{\cS}{\mathcal{S}}

\newcommand{\cX}{\mathcal{X}}

\newcommand{\cY}{\mathcal{Y}}

\newcommand{\cV}{\mathcal{V}}
\newcommand{\cU}{\mathcal{U}}

\def\argmax{\operatorname*{argmax\,}}

\def\exp{\operatorname*{exp\,}}
\def\log{\operatorname*{log\,}}

\def\btheta{{\boldsymbol \theta}}

\def\x{{\boldsymbol x}}

\def\y{{\boldsymbol y}}

\def\gradient{{\nabla}}
\def\expect{{\mathbb E}}

\def\Prr{{\displaystyle P}}

\newcommand{\comment}[1]{}

\newcommand{\fnorm}[2][2]{\ensuremath{ \left\| #2 \right\|^2_{ \mathrm{#1} } } }

\def\x{{\boldsymbol x}}

\newcommand*\rot{\rotatebox{90}}

\def\best{\bf \cellcolor[gray]{0.85}}

%% file: intro.tex
\section{Introduction}\label{sec:introduction}

Semantic image segmentation aims to predict a category label for every image pixel,
which is an important yet challenging task for image understanding.
Recent approaches have applied convolutional neural network (CNNs) \cite{farabet2013learning,LongSD14,ChenPKMY14}
to this pixel-level labeling task and achieved remarkable success.
Among these CNN-based methods, fully convolutional neural networks (FCNNs)~\cite{LongSD14,ChenPKMY14}
have become a popular choice, because of their computational efficiency for dense prediction and end-to-end style learning.

Contextual relationships are ubiquitous and provide important cues for scene understanding tasks.
Spatial context can be formulated in terms of  semantic compatibility relations between one object and its neighboring objects or image patches (stuff), in which a compatibility relation is an indication of the co-occurrence of visual patterns.
For example, a car is likely to appear over a road, and a glass is likely to appear over a table.
Context can also encode incompatibility relations.
For example, a boat is unlikely to appear on a road.  These relations also exist
at finer scales, for example, in
object part-to-part relations, and part-to-object relations.
In some cases, contextual information is the most important cue, particularly when a single object shows significant visual ambiguities.
A more detailed discussion of the value of spatial context can be found in \cite{heitz2008learning}.

\begin{figure*}[t]
	\centering
	\includegraphics[width=0.85\linewidth]{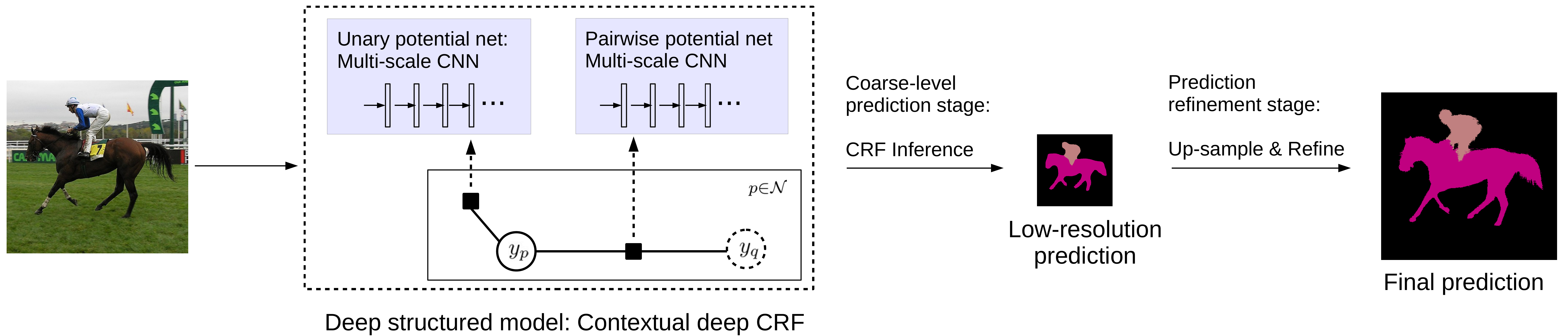}
\caption{An illustration of the prediction process of our method.
Both our unary and pairwise potentials are formulated as multi-scale CNNs for capturing semantic relations between image regions.
Our method outputs low-resolution prediction after performing CRF inference, then the prediction is up-sampled and refined in a standard post-processing stage to output the final prediction.}
\label{fig:general_graph}
\end{figure*}

We explore two types of spatial context to improve the segmentation performance: patch-patch context and patch-background context.
The patch-patch context is the semantic relation between the visual patterns of two image patches. Likewise,
patch-background context the semantic relation between an image patch and a large background region.

Explicitly modeling the patch-patch contextual relations has not been well studied in recent CNN-based segmentation methods.
In this work, we propose to explicitly model the contextual relations using conditional random fields (CRFs).
We formulate CNN-based pairwise potential functions to capture semantic correlations between neighboring patches.
Some recent methods  combine CNNs and CRFs for semantic segmentation,
e.g., the dense CRFs applied in \cite{ChenPKMY14,schwing2015fully,zheng2015conditional,Dai2015arXiv}.
The purpose of applying the dense CRFs in these methods is to refine the upsampled low-resolution prediction to sharpen object/region boundaries.
These methods consider Potts-model-based pairwise potentials for enforcing local smoothness.
There the pairwise potentials are conventional log-linear functions.
In contrast,
here we learn more general pairwise potentials using CNNs to model the semantic compatibility between image regions.
Our CNN pairwise potentials aim to improve the coarse-level prediction rather than  merely   encouraging   local smoothness,
and thus have a different purpose compared to Potts-model-based pairwise potentials.
Given that these two types of potentials make different effects, they can be combined to improve  segmentation results.
Fig.~\ref{fig:general_graph} illustrates the prediction process of our method.

In contrast to patch-patch context,
patch-background context is widely explored in the literature.
For CNN-based methods,
background information can be effectively captured
by combining features from a multi-scale image network input, and has shown good performance
in some recent segmentation methods \cite{farabet2013learning,MostajabiYS14}.
A special case of capturing patch-background context is considering the whole image as the background region and incorporating the image-level label information into learning.
In our approach, to encode rich background information, we construct multi-scale networks and apply sliding pyramid pooling on feature maps.
The traditional pyramid pooling (in a sliding manner) on the feature map is able to capture information from background regions of different sizes.

Incorporating general pairwise potentials usually involves computationally expensive inference, which brings challenges for CRF learning.
To facilitate efficient learning we apply piecewise training of the CRF \cite{SuttonM05} to avoid repeated inference
 during back propagation training of the deep model.

Thus our main contributions are as follows.
\begin{itemize}
\item We formulate CNN-based general pairwise potential functions in CRFs to explicitly model patch-patch semantic relations.

\item
  Deep CNN-based general pairwise potentials are challenging for efficient CNN-CRF joint learning.
  We perform approximate training, using piecewise training of CRFs \cite{SuttonM05},
  to avoid the repeated inference at every stochastic gradient descent iteration and thus achieve efficient learning.

\item
  We explore background context by applying a network architecture with traditional multi-scale image input \cite{farabet2013learning} and sliding pyramid pooling
  \cite{lazebnik2006beyond}. We empirically demonstrate the effectiveness of this network architecture for semantic segmentation.

\item
  We set new state-of-the-art performance on a number of challenging semantic segmentation datasets,
  including NYUDv2, PASCAL VOC 2012, PASCAL-Context, SIFT-flow, SUN-RGBD, Cityscapes dataset and so on.
	In particular, we achieve an intersection-over-union score of $77.8$ on the PASCAL VOC 2012 dataset.
\end{itemize}

\begin{figure*}[t]
  \center
  \includegraphics[width=.99999\linewidth]{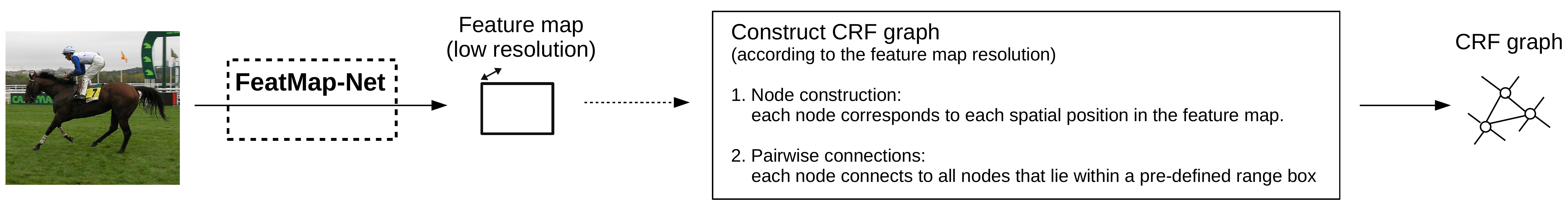}
\caption{An illustration of generating a feature map with FeatMap-Net and constructing the CRF graph.}
\label{fig:crf_graph}
\end{figure*}

\subsection{Related work}
\label{sec:relatedwork}

Preliminary results of our work appeared in \cite{lin2016piece}.
Exploiting contextual information
has been widely studied in the literature (e.g., \cite{rabinovich2007objects,heitz2008learning,doersch2014context}).
For example,  early work of ``TAS'' \cite{heitz2008learning} models different types of spatial context between {\em Things} and {\em Stuff}
using a generative graphical model.

The most successful recent methods for semantic image segmentation are based on CNNs. 
CNN based methods have shown outstanding performance compared to traditional semantic segmentation methods like TextonBoost \cite{shotton2009textonboost}.
A number of these CNN-based methods are region-proposal-based methods \cite{GirshickDDM13,BharathECCV2014}, which first generate region proposals and then assign category labels to each of them.
Very recently,  FCNNs
 \cite{LongSD14,ChenPKMY14,Dai2015arXiv} have become a popular choice for their efficient feature generation and end-to-end training.
FCNNs have also been applied to a range of other dense-prediction tasks recently,
 such as image restoration \cite{Eigen_iccv13}, image super-resolution \cite{Dong_eccv14}
 and depth estimation \cite{dcnn_nips14,liu2014deep,eigen2015predicting}.
The method that we propose here is  also  built upon fully convolution-style networks.

FCNN methods make use of the Image-Net trained CNN models
(e.g., the VGG-16 model \cite{simonyan2014very}) which takes advantages of the large Image-Net dataset for learning deep models.
For convolution and pooling layers, the resolution of the output feature map is down-sampled if
the convolution/pooling stride is greater than $1$.
Usually a few such layers use a stride setting of $2$,
hence the direct predictions of FCNNs are typically in low resolution.
To increase the prediction resolution,
the naive  method  of directly reducing the strides for all layers is not able to address this down-sampled prediction for a deep network.
Small strides result in prohibitively expensive computation for a deep network, and also reduce the view-of-field (the image region that a filter is able to ``see") of the network layers.
Network layers with insufficient view-of-field may not be able to capture high-level semantic patterns and thus degrade the performance.

To address this low-resolution prediction issue,
a variety of FCNN based methods are proposed very recently which focus on refining the low-resolution prediction to obtain high resolution prediction.
DeepLab-CRF \cite{ChenPKMY14} first applies atrous convolution to produce larger size feature maps and performs bilinear upsampling on the prediction score map to the input image size, then they apply the dense CRF method \cite{krahenbuhl2012efficient} to refine the object boundary by levering  low-level (color contrast) information.
They consider Potts-model based pairwise potential functions which enforce local smoothness.
CRF-RNN \cite{zheng2015conditional} extends this approach by implementing the mean field CRF inference as recurrent layers for end-to-end learning of the dense CRF and  FCNN network.
The work in \cite{noh2015learning} learns deconvolution layers to upsample the low-resolution predictions.
The depth estimation method \cite{liu2015learning} explores super-pixel pooling
for building the gap between the low-resolution feature map and high-resolution final prediction.
Eigen \textit{et~al.} \cite{eigen2015predicting} perform coarse-to-fine learning of multiple networks with different resolution outputs for refining the coarse prediction.
The method FCN \cite{LongSD14} and Hyper-column \cite{hariharan2014hypercolumns} explore mid-layer features (skip connections) for high-resolution prediction.
Unlike these methods, our method focuses on improving the coarse (low-resolution) prediction
by learning general CNN pairwise potentials to capture semantic relations between patches.
These  methods are complementary to our method.

Jointly learning CNNs and CRFs has also been explored in other applications apart from segmentation.
Recent work in \cite{liu2014deep,liu2015learning} proposes to jointly learn {\em continuous} CRFs and CNNs for depth estimation
from  single monocular images. They focus on continuously-valued variable prediction,
while our method is for discrete categorical label prediction.
The work in \cite{Lecun_nips14} combines CRFs and CNNs for human pose estimation.
The authors of \cite{chen2014learning} explore joint training of Markov random fields and deep neural networks for the tasks of predicting words from noisy images and  multi-class classification.
They require marginal inference for every gradient calculation which is computationally expensive for training deep models.

%% file: method.tex
\begin{figure}[t]
  \center
  \includegraphics[width=.85\linewidth]{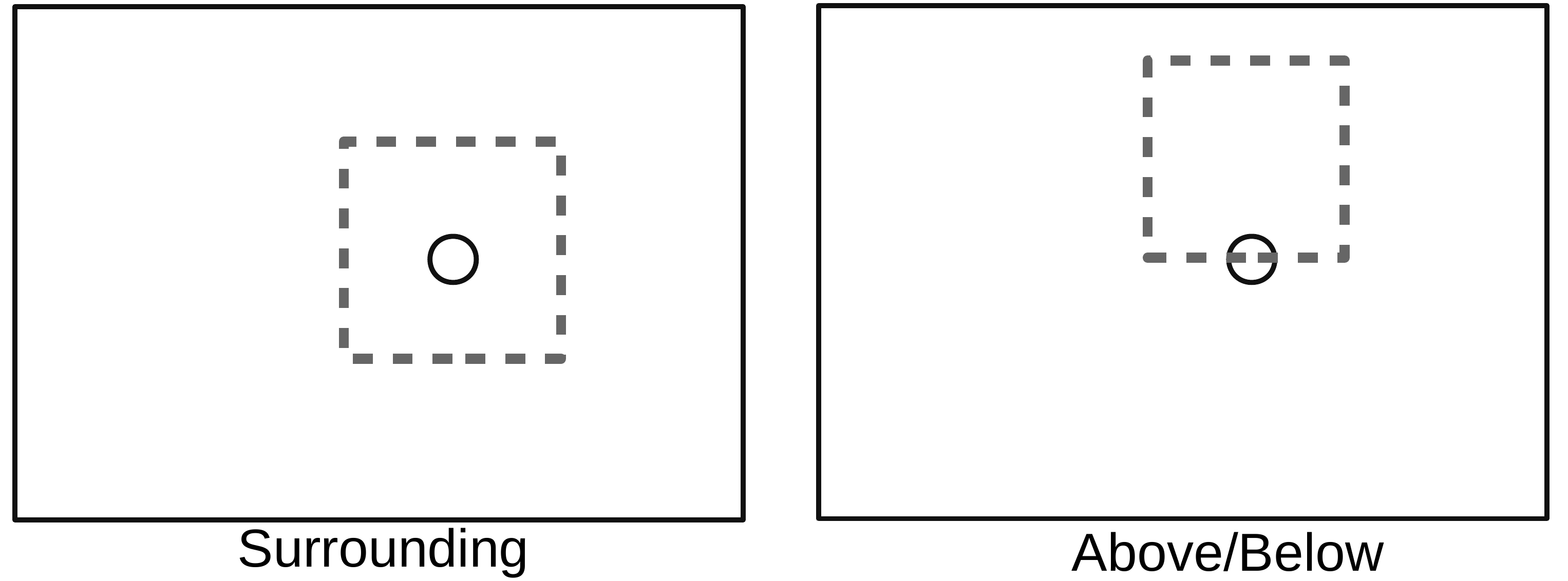}
\caption{An illustration of constructing pairwise connections in a CRF graph.
A node is connected to all other nodes which lie inside the range box (dashed box in the figure).
Two types of spatial relations are described in the figure, which
correspond to two types of pairwise potential functions.
}
\label{fig:pairwise}
\end{figure}

\section{Modeling semantic pairwise relations}

We first describe how to build the CRF graph for modeling semantic pairwise relations.
Given an image, we first apply a convolutional network to generate a feature map.
We refer to this network as `FeatMap-Net', details of which are presented in Sec.~\ref{sec:network_details}
(Fig.~\ref{fig:featmapnet} shows the overall architecture).
With this feature map, we construct one node in the CRF graph corresponding to one spatial position of the feature map.
Fig.~\ref{fig:crf_graph} illustrates how we construct nodes and pairwise connections in a CRF graph.

Pairwise connections are constructed by connecting one node to all other nodes which lie within a spatial range box (the dashed box in Fig.~\ref{fig:pairwise}).
We consider different spatial relations by defining different types range boxes, and each type of spatial relation is modeled by a specific pairwise potential function.
As shown in Fig.~\ref{fig:pairwise},  our method models the ``surrounding" and ``above/below" spatial relations.
For the surrounding relation, the range box is centered at the node. For the above/below relation, the
bottom edge of the range box is centered at the node.

In our experiments, the size of the range box (dash box in the figure) size is $0.4a \times 0.4a$,
where $a$ is the length of the short edge of the feature map.
It would be straightforward to construct more pairwise potentials, by varying either the sizes or positions of the connection range boxes, and our approach is not limited to connections within ``boxes''.

\section{Contextual Deep CRFs}
\label{sec:crf_details}

Here we present the details of our deep CRF model.
We denote by $\x \in \cX$ one input image and $\y \in \cY$ the labeling mask which describes the label configuration of each node in the CRF graph.
The energy function is denoted by $E(\y, \x; \btheta)$ which models the compatibility of the input-output pair, with a small output value indicating high confidence in the prediction $\y$.
All network parameters are denoted by $\btheta$ which we need to learn.
The conditional likelihood for one image is formulated as follows:
\begin{equation}\label{eq:prob}
\begin{aligned}
\Prr(\y|\x) = \frac{1}{Z(\x)} \exp [- E(\y, \x)].
\end{aligned}
\end{equation}
Here $Z$ is the partition function, defined as:
$Z(\x) = \sum_{\y} \exp [ -E(\y, \x) ]$.
The energy function is typically formulated by a set of unary and pairwise potentials:
\begin{align}
\label{eq:energy}
	E(\y, \x) = &  \sum_{U \in \cU} \sum_{p \in \cN_U } U(y_{p}, \x_p) \notag \\
  & + \sum_{V \in \cV} \sum_{(p,q) \in \cS_V} V(y_{p}, y_{q}, \x_{pq}). 
\end{align}
Here $U$ is a unary potential function.
To make the exposition more general, we consider multiple types of unary potentials with
$\cU$ the set of all such unary potentials.
$\cN_U$ is a set of nodes for the potential $U$.
Likewise, $V$ is a pairwise potential function
with $\cV$ the set of all types of pairwise potential. $\cS_V$ is the set of edges for the potential $V$.
$\x_{p}$ and $\x_{pq}$  indicates the corresponding image regions which associate to the specified node and edge.

\begin{figure*}[t]
	\centering
	\includegraphics[width=.9\linewidth]{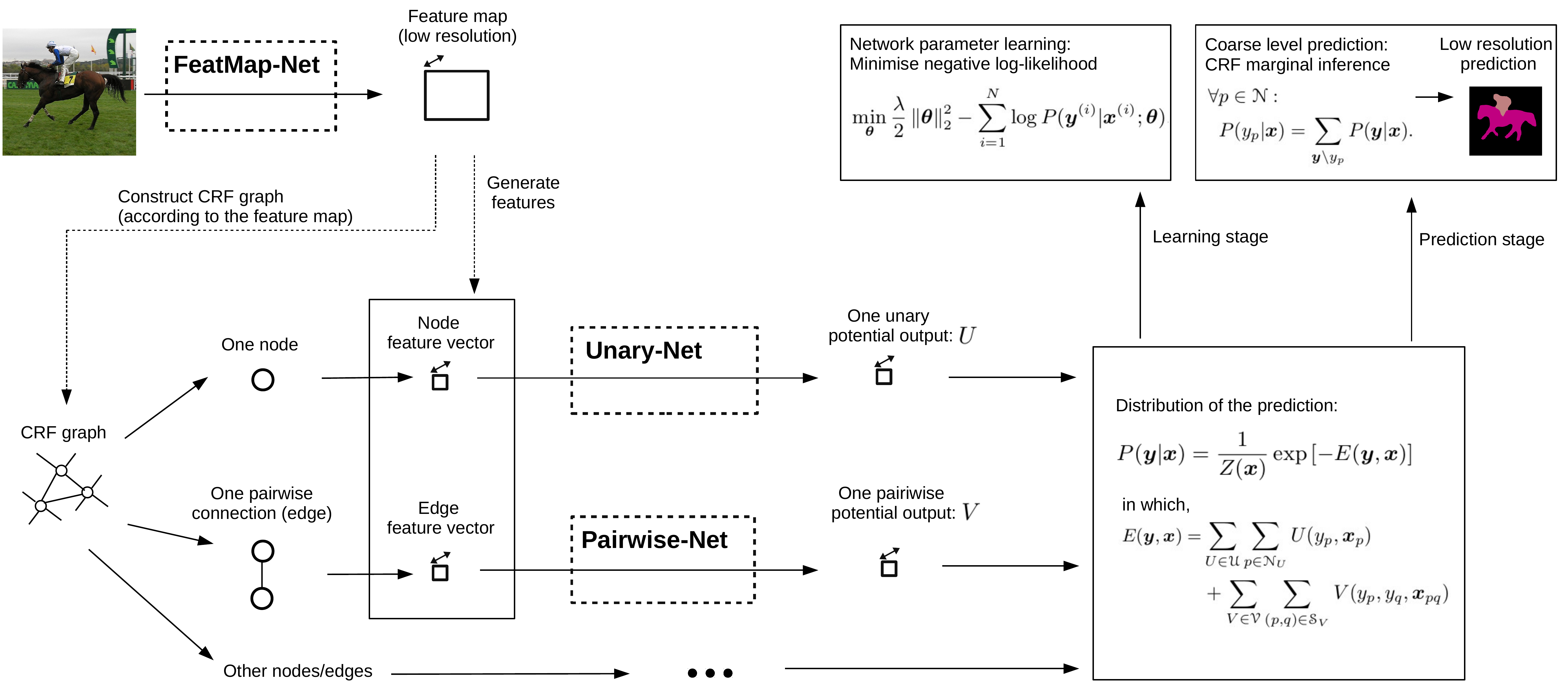}
\caption{An overview of the proposed contextual deep structured model.
Unary-Net and Pairwise-Net are shown here for generating potential function outputs.}
\label{fig:potential_net}
\end{figure*}

The potential function is constructed by a deep network for generating feature map (FeatMap-Net) and a shallow network (Unary-Net or Pairwise-Net) to generate the output of the potential function. Details are described in the following sections.
An overview of our contextual deep structured model for prediction and training is shown in Fig.~\ref{fig:potential_net}.

\begin{figure*}[t]
	\centering
	\includegraphics[width=.9\linewidth]{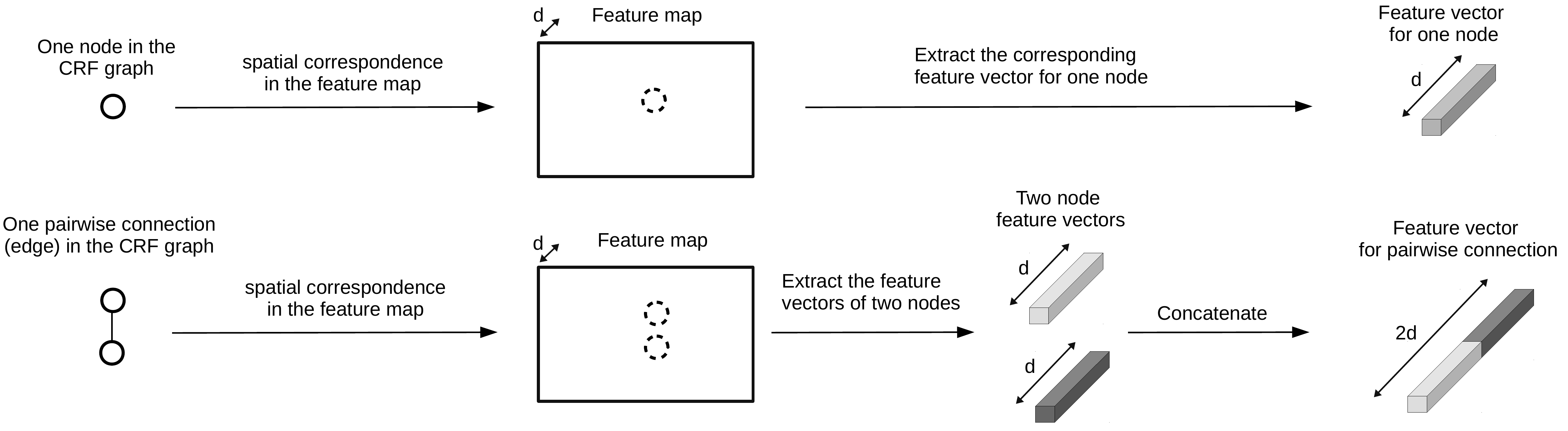}
\caption{An illustration of generating feature vectors
for CRF nodes and pairwise connections from the feature map output by FeatMap-Net.
The symbol $d$ denotes the feature dimension. We concatenate the corresponding features of two connected nodes in the feature map to obtain the CRF edge features.}
\label{fig:node_feat}
\end{figure*}

\subsection{Unary potential functions}
We formulate the unary potential function by stacking the FeatMap-Net for generating feature maps and
a shallower fully connected network (referred to as Unary-Net) to generate the final output of the unary potential function.
The unary potential function is written as follows:
\begin{equation}
	U(y_{p}, \x_p; \btheta_U) = - z_{p,y_p} (\x; \btheta_U).
\end{equation}
Here $z_{p, y_p }$ is the output value of Unary-Net, which corresponds to the $p$-th node and the $y_p$-th class.

Fig.~\ref{fig:potential_net} shows an illustration of the Unary-Net and how it corporates with FeatMap-Net.
Fig.~\ref{fig:node_feat} demonstrates the process for generating the feature vector for one node.
The input of the Unary-Net is the node feature vector extracted from the feature map which is generated by FeatMap-Net.
The feature vector for one CRF node is simply the corresponding feature vector in the feature map.
The dimension of the Unary-Net output vector for one node is $K$,
which is the same as the number of classes.

\begin{figure*}[t]
	\center
	\includegraphics[width=1\linewidth]{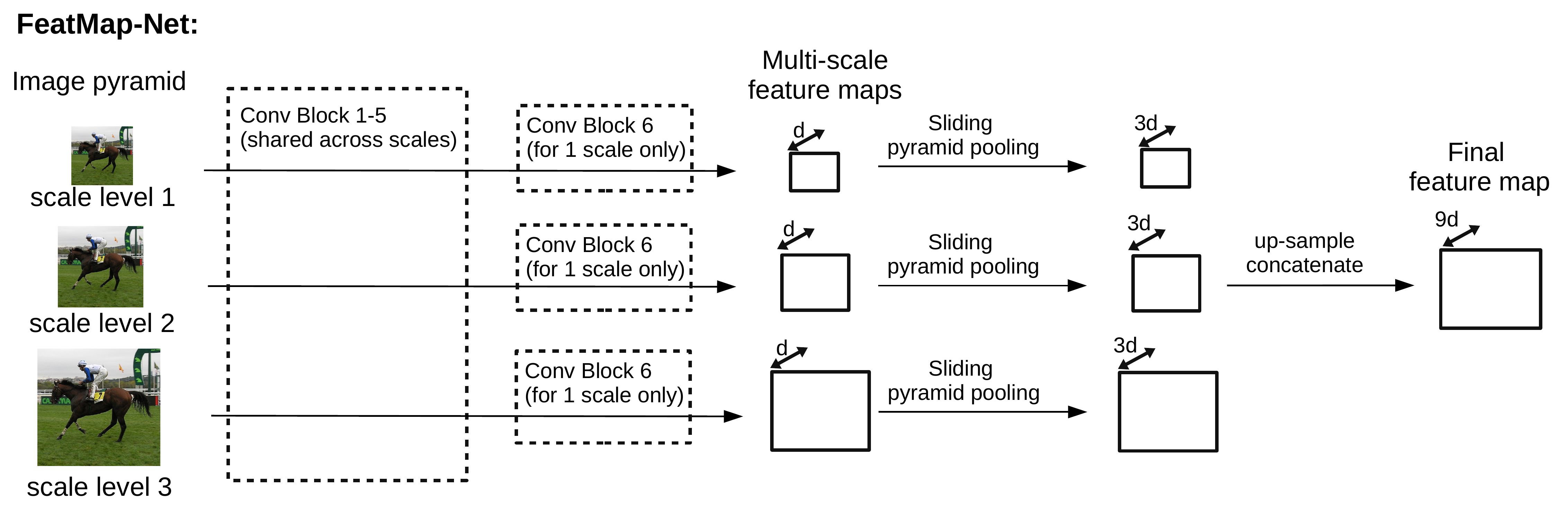}
\caption{The details of our FeatMap-Net.
An input image is first resized into $3$ scales,
then each resized image goes through 6 convolution blocks to output one feature map.
Top $5$ convolution blocks are shared for all scales. Every scale has a specific convolution block (Conv Block 6).
We perform $2$-level sliding pyramid pooling and concatenate the pooled feature map to the original feature map.
The symbol $d$ denotes the feature dimension.
}
\label{fig:featmapnet}
\end{figure*}

\subsection{Pairwise potential functions}
We formulate the unary potential function, analogous to the unary potentials,
by stacking the FeatMap-Net for generating feature maps and
a shallower fully connected network (referred to as Pairwise-Net) to generate the final output of the pairwise potential function.
The pairwise potential function is written as follows:
\begin{align}
	V & (y_{p}, y_{q},  \x_{pq}; \btheta_V) = - z_{p, q, y_p, y_q} (\x; \btheta_V).
\end{align}
Here $z_{p, q,  y_p,  y_q }$ is the output value of Pairwise-Net.
It is the confidence value for the node pair $(p, q)$ when they are labeled with the class value $( y_p,  y_q )$, which measures the compatibility of the label pair $(y_{p}, y_{q}$) given the input image $\x$.
$\btheta_V$ is the corresponding set of CNN parameters for the potential $V$, which we need to learn.
The role of Pairwise-Net in our structured model is illustrated in Fig.~\ref{fig:potential_net}. 
Fig.~\ref{fig:node_feat} describes the process for generating the feature vector for one pairwise connection.
The input of Pairwise-Net is the edge feature vector which is generated from the feature map for two connected nodes.
Following the work of~\cite{kolesnikov2014closed},
we concatenate the corresponding feature vectors of two connected nodes
to obtain the CRF edge feature vector.
The Pairwise-Net has $K^2$ output units to match the number of possible label combinations for a pair of nodes.

Our formulation of pairwise potentials is different from the Potts-model-based
smoothness potentials in the existing methods of \cite{ChenPKMY14,zheng2015conditional}.
The Potts-model-based pairwise potentials are a log-linear functions and
employ a special formulation for enforcing
neighborhood smoothness based on color contrast, and thus to sharpen object/region boundaries.
In contrast,
our pairwise potentials model the semantic compatibility relations between two nodes
with the output for every possible value of the label pair $(y_{p}, y_{q})$
individually parameterized by CNNs.
Clearly, these two types of pairwise potential formulations have different
purposes and effects.

Most recent segmentation methods, e.g., the work in \cite{ChenPKMY14,Dai2015arXiv,schwing2015fully,zheng2015conditional},
have applied the dense CRF method \cite{krahenbuhl2012efficient} in the prediction refinement stage
for refining (sharpen object boundaries) the coarse (low-resolution) prediction.
The dense CRF method is a Potts-model-based fully-connected CRF with
 pairwise potentials based on color contrast  for local smoothness.
It is important to clarify that, this smoothness CRFs and our contextual deep CRFs are working in different prediction stages.
Our contextual CNN pairwise potentials are applied in the coarse prediction stage to improve the lower-resolution prediction,
rather than applying in the boundary refinement stage.

In our framework, after obtaining the coarse level prediction, we still
 need to perform a refinement step to obtain the final high-resolution prediction
(as shown in Fig.~\ref{fig:general_graph}).
Hence we also apply the dense CRF method \cite{krahenbuhl2012efficient}, as in many other recent methods,
in the prediction refinement step.
Therefore, our method takes advantage of both contextual CNN potentials and the traditional smoothness potentials to improve the final result.
More details for prediction can be found in Sec.~\ref{sec:prediction}.

\subsubsection{Asymmetric pairwise potentials}
As in \cite{winn2006layout,heesch2010markov},
modeling asymmetric relations requires learning asymmetric potential functions,
the output of which  should depend on the input order of a pair of nodes.
In other words, the potential function is required to be capable of modeling different input orders.
Typically we have the following case for asymmetric relations:
\begin{align}
	V & (y_{p}, y_{q},  \x_{pq}) \neq V (y_{q}, y_{p},  \x_{qp}).
\end{align}
Ideally, the potential $V$ is learned from the training data.

Here we discuss the asymmetric relation ``above/below" as an example.
We take advantage of the input pair order to indicate the spatial configuration of two nodes,
thus the input $(y_p, y_q, \x_{pq})$
indicates the configuration that the node $p$ is spatially lies above the node $q$.
Clearly, the potential function is required to model different input orders.

The asymmetric property is readily achieved with our general formulation of pairwise potentials.  The edge features for the node pair $(p, q)$ are generated from a concatenation of the corresponding features of nodes $p$ and $q$ (as in \cite{kolesnikov2014closed}), in that order.  The potential output for every possible pairwise label combination for $(p,q)$ is individually parameterized by the pairwise CNNs.  These factors ensure that the edge response is order dependent, easily satisfying the asymmetric requirement.

\begin{figure}[t]
	\center
	\includegraphics[width=1\linewidth]{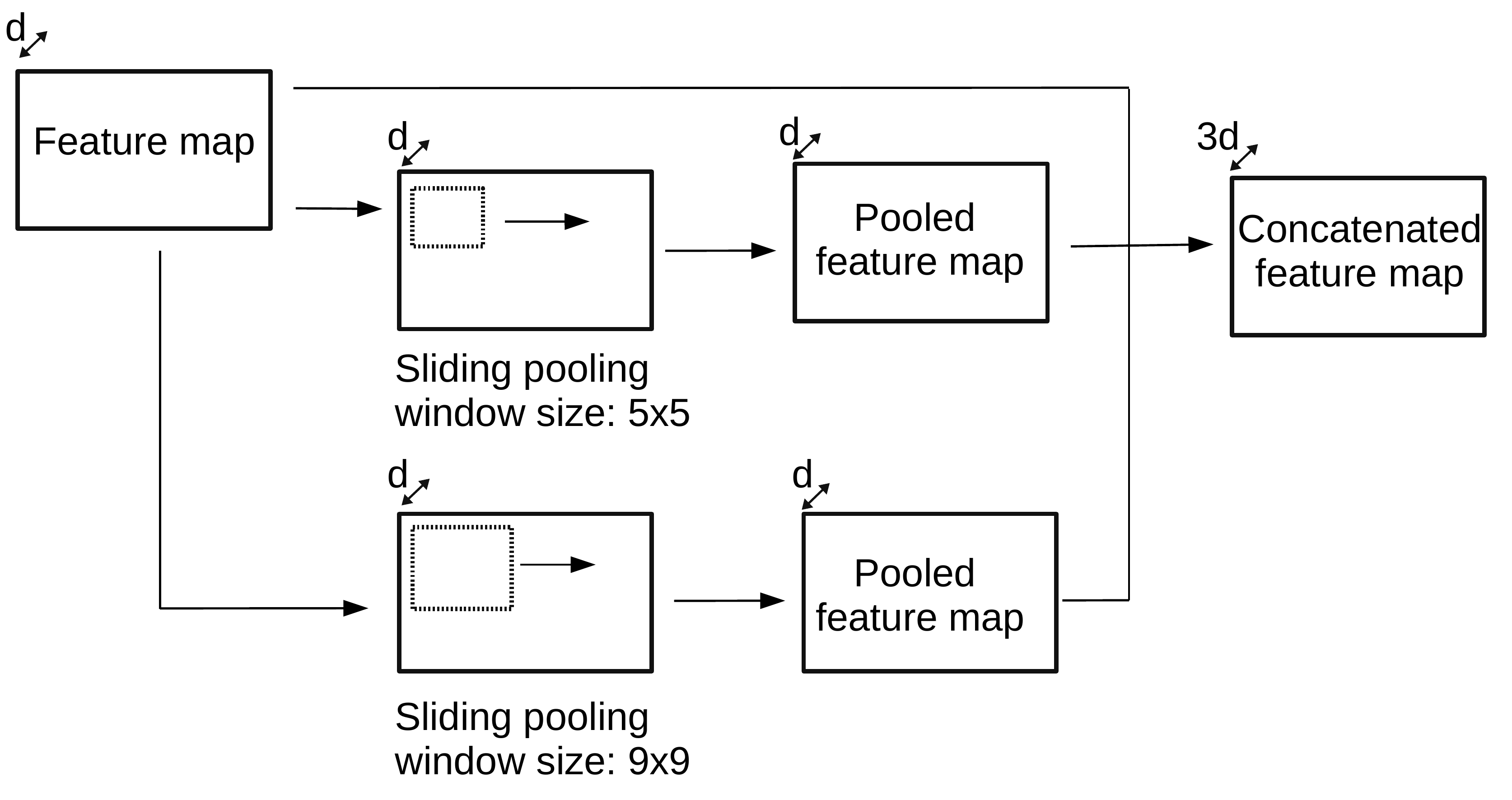}
\caption{
Details for sliding pyramid pooling.
We perform $2$-level sliding pyramid pooling on the feature map for capturing patch-background context,
which encode rich background information and increase the field-of-view for the feature map.
}
\label{fig:pooling}
\end{figure}

\section{Exploring background context}
\label{sec:network_details}

We develop multi-scale CNNs and sliding pyramid pooling in our FeatMap-Net to encode rich background information for capturing patch-background context.
Fig.~\ref{fig:featmapnet} shows the architecture of FeatMap-Net.
Details are  presented shortly in the squeal.

Applying CNNs on multi-scale images has shown improved performance in some recent segmentation methods, e.g., \cite{farabet2013learning,MostajabiYS14}.
In our multi-scale network, an input image is first resized into $3$ scales,
then each resized image goes through 6 convolution blocks to output one feature map.
In our experiment, the $3$ scales for the input image are set to $1.2$, $0.8$ and $0.4$.
All scales share the same top $5$ convolution blocks.
In addition, each scale has an exclusive convolution block (``Conv Block 6" in the figure)
which captures scale-dependent information.
The resulting $3$ feature maps (corresponding to $3$ scales) are of different resolutions,
 therefore we  upscale the two smaller ones to the size of the largest feature map using bilinear interpolation.
 These feature maps are then concatenated to form one feature map.

We perform spatial pyramid pooling \cite{lazebnik2006beyond} (a modified version using sliding windows)
on the feature map to capture information from background regions in multiple sizes.
From another point of view, this increases the field-of-view for the feature map, for which
 feature vectors are able to encode information from a larger image region.
Increasing the field-of-view generally helps to improve performance,
which is also discussed in \cite{ChenPKMY14}.

The details of spatial pyramid pooling are illustrated in Fig. \ref{fig:pooling}.
In our experiment, we perform $2$-level pooling for each image scale.
We define $5 \times 5$ and $9 \times 9$ sliding pooling windows with max-pooling to generate $2$ sets of pooled feature maps.
These pooled feature maps are then concatenated to the original feature map to construct the final feature map, and thus the resulting feature dimension is $512 \times 3$ for one image scale.

\section{Network configurations}
\label{sec:netconfig}

\begin{figure}[t]
	\center
	\includegraphics[width=.75\linewidth]{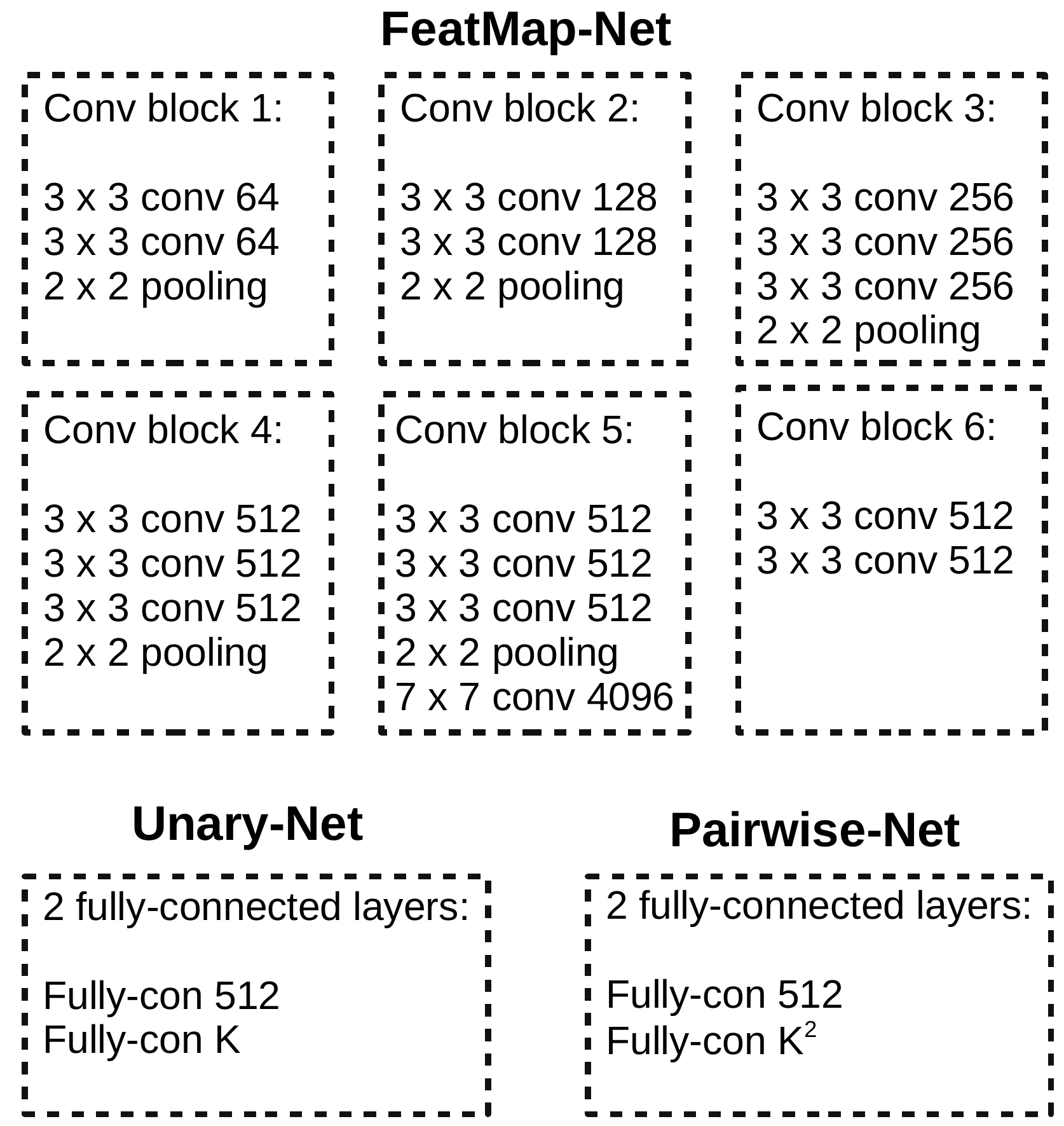}
\caption{The detailed configuration of the networks: FeatMap-Net, Unary-Net and Pairwise-Net.
  $K$ is the number of classes.
  The filter size for convolution and the number of filters are shown for all layers.
  For FeatMap-Net,   the top $5$ convolution blocks share the same configuration
as the convolution blocks in the VGG-16 network. The stride of the last max pooling layer is 1, and
for the other max pooling layers we use the same stride setting as the VGG-16 network.
}
\label{fig:network_conf}
\end{figure}

We show the detailed network layer configuration for all networks in Fig.~\ref{fig:network_conf}.
For FeatMap-Net, the configuration of the convolution blocks is similar to the VGG-16 model \cite{simonyan2014very}.
The top $5$ convolution blocks share the same configuration as the VGG-16 network.
The first fully-connected layer in VGG-16 is converted into a convolution layer ( see FCN in \cite{LongSD14} for details) and  merged into the $5$-th convolution block.
We only transfer the first fully-connected (FC) layer into our network rather than $2$ FC layers.
Note that transferring $2$ FC layers is commonly applied in almost all recent FCN based methods \cite{LongSD14,ChenPKMY14,zheng2015conditional}.
The FC layer in the VGG-16 model contains a large number of filters ($4096$),
thus our network which transfers only one FC layer is more efficient.

In FeatMap-Net, we add a new convolution block (``Conv Block 6" in the figure) which contains $2$ convolution layers.
This extra convolution block is not existed in the VGG-16 network.
With this new convolution block, we are able to capture scale-dependent information and increase the abstraction level.
We also have the consideration of increasing the field-of-view for the final feature map by adding this extra block.

As discussed in Sec.~\ref{sec:relatedwork},
The stride setting of the convolution and pooling layers will result in a feature map which has a smaller resolution than the input image.
For the convolution and pooling layers, the resolution of the output feature map is down-sampled if
the stride is greater than $1$.
Note that there are a number of convolution/pooling layers in VGG-16 model which use the stride setting of $2$.
Therefore,
for the original VGG-16 model, the resolution of the output feature map is $32$ times smaller than the size of the input image (see FCN \cite{LongSD14} for details).
To increase the resolution of the feature map, almost all recent VGG-16 based methods \cite{LongSD14,ChenPKMY14,zheng2015conditional}
reduce the stride of the last two pooling layers to $1$, which reduces the down-sampling factor from $32$ to $8$.

In our setting, we reduce the stride of the last max pooling layer (only one layer) in the VGG-16 network to $1$, instead of reducing for two pooling layers in many other methods \cite{ChenPKMY14,zheng2015conditional}. The resolution of the resulting feature map is $16$ times smaller than the size of the input image.
For a $500 \times 500$ input image, the resolution of the resulting feature map is around $30 \times 30$.

Directly changing the stride inevitably  degrades the performance of the learned filters
since the field-of-view of the input feature map for some filters is changed.
To preserve the field-of-view, recent work has proposed a number of approaches.
For example, a straightforward approach is to increase the receptive field size of the filter
(e.g., double the filter size). Large filter sizes will significantly increase the computation cost
 for convolution operations.
This approach also brings the problem of how to upsample the filter weights.
Probably a better approach is to apply the hole algorithm as in \cite{ChenPKMY14},
which performs a skipping (sampled) dot-product calculation for filter convolution.
Therefore, a large convolution window size can be applied without increasing the computation cost.

Different from existing approaches, here we apply a simple yet effective approach.
We add extra two $3 \times 3$ convolution layers (``Conv Block 6" ) instead of increasing the filter size.
These extra layers are able to enlarge the field-of-view and compensate the side-effect of reducing the stride in pooling layers.

\begin{figure}[t]
	\center
	\includegraphics[width=.9\linewidth]{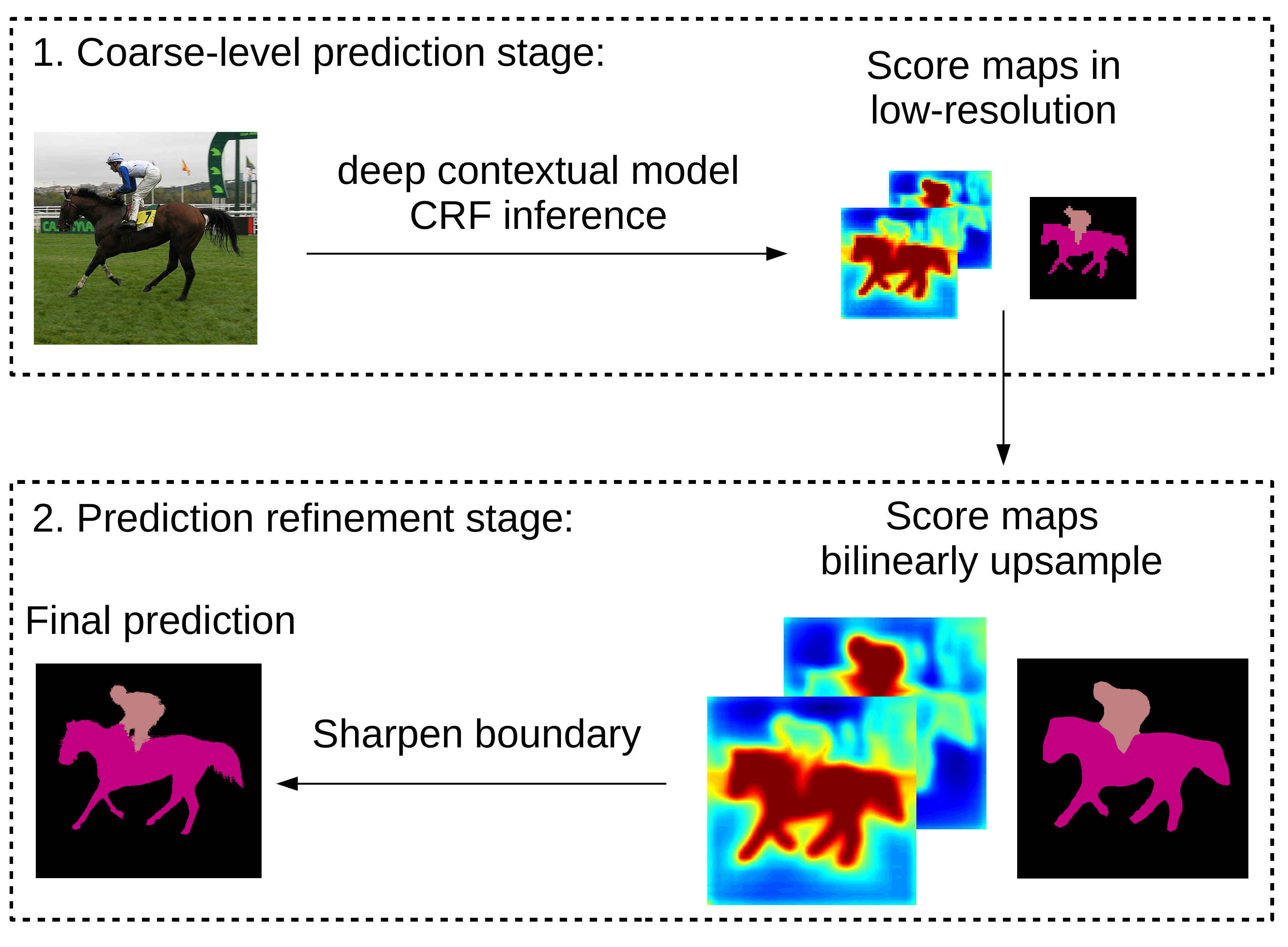}
\caption{
An illustration of our two-stage prediction process.
The prediction process consists of two stages:
the coarse-level prediction stage and the prediction refinement stage.
We first perform CRF inference on our contextual model to generate a score map for coarse-level prediction, then we bilinearly unsmaple the score map
and apply a boundary refinement method \cite{krahenbuhl2012efficient} to obtain
the final prediction which has the same resolution as the input image.
}
\label{fig:prediction_stage}
\end{figure}

\section{Prediction}
\label{sec:prediction}

At the prediction stage,
our deep structured model  generates low-resolution prediction (as shown in Fig.~\ref{fig:general_graph}),
which is $1/16$ of the input image size.
As discussed in Sec. \ref{sec:netconfig}, this is due to the stride setting of pooling layers.
Therefore, we apply two prediction stages for obtaining the final high-resolution prediction:
the coarse-level prediction stage and the prediction refinement stage.
We first perform CRF inference on our contextual structured model to generate a score map for coarse-level prediction, then we bilinearly unsmaple the score map
and apply a boundary refinement method \cite{krahenbuhl2012efficient} to obtain the final prediction which has the same resolution as the input image.
This two-stage prediction process is illustrated in Fig.~\ref{fig:prediction_stage}.

\subsection{Coarse-level prediction stage}

We perform CRF inference on our contextual structured model to obtain the coarse prediction of a test image.
For example, we can solve the maximum a posteriori (MAP) problem: $\y^{\star}=\argmax_{\y} \Prr(\y|\x)$.
Alternatively, we also can consider the marginal inference over nodes for prediction:
\begin{align} \label{eq:inference}
\forall p \in \cN: \;\; \Prr(y_p|\x)&= \sum_{\y \backslash y_p } \Prr(\y|\x).
\end{align}
We obtain the marginal distribution for each node after performing this marginal inference.
This marginal distribution can be further applied in the next prediction stage for boundary refinement.
Details are shown
 in the next section.

Our CRF graph does not form a tree structure,
nor are the potentials submodular,
hence we need to an apply approximate inference.
To address this we
apply an efficient  message passing algorithm which is based on the mean field approximation \cite{Nowozinstruct}.
The mean field algorithm constructs a simpler distribution $Q(\y)$, e.g.,
a product of independent marginals:
$Q(\y)=\prod_{p \in \cN} Q_p (y_p)$,
which minimizes the KL-divergence between the distribution $Q(\y)$ and $\Prr(\y)$.
In our experiments, we perform $3$ mean field iterations.

\subsection{Prediction refinement stage}
We generate the score map for the coarse prediction from the marginal distribution
which we obtain from the mean-field inference.
We first bilinearly up-sample the score map of the coarse prediction to the size of the input image.
Then we apply a common post-processing method \cite{krahenbuhl2012efficient} (dense CRF)
to sharpen the object boundary for generating the final high-resolution prediction.
This post-processing method leverages low-level pixel intensity information (color contrast) for boundary refinement.
Note that most recent work on image segmentation produce low-resolution prediction and have a upsampling and refinement process/model for the final prediction, e.g.,
\cite{ChenPKMY14,zheng2015conditional,Dai2015arXiv}.

In summary, we simply perform bilinear upsampling of the coarse score map and apply the boundary refinement post-processing.
We argue that this stage can be further improved by applying more sophisticated refinement methods, e.g., training deconvolution networks \cite{noh2015learning}
training multiple coarse to fine learning networks \cite{eigen2015predicting},
and exploring middle layer features for high-resolution prediction \cite{hariharan2014hypercolumns,LongSD14}.
It is expected that applying better refinement approaches will gain further performance improvement.
In the experiment part, we show an example of exploring the feature maps from middle layers to refine the coarse prediction.
We apply this improved refinement approach on the dataset PASCAL VOC 2012.
Refer to Sec.~\ref{sec:voc2012exp} for details.

\section{CRF training}

A common approach to CRF learning is to maximize the likelihood,
or equivalently minimize the negative log-likelihood, which can be written
for one image as:
\begin{align}
 - \log \Prr(\y | \x; \btheta) = E(\y, \x; \btheta) + \log Z(\x; \btheta).
\label{eq:Prr}
\end{align}
Adding regularization to the CNN parameter $\btheta$, the optimization problem for CRF learning is: 
\begin{align}
\min_{\btheta} \frac{\lambda}{2} \fnorm \btheta -\sum_{i=1}^N \log \Prr(\y^{(i)} | \x^{(i)}; \btheta).
\label{eq:opt}
\end{align}
Here $\x^{(i)}$, $\y^{(i)}$ denote the $i$-th training image and its segmentation mask; $N$ is the number of training images; $\lambda$ is the weight decay parameter.
Substituting \eqref{eq:Prr} into \eqref{eq:opt} yields:
\begin{align}
\label{eq:crf_learning_org}
\min_{\btheta} \frac{\lambda}{2} \fnorm \btheta + \sum_{i=1}^N \biggr[ E(\y^{(i)}, \x^{(i)}; \btheta) + \log Z(\x^{(i)}; \btheta) \biggr].
\end{align}
We can apply stochastic gradient (SGD) based methods to optimize the above problem for learning $\btheta$.
The energy function $E (\y, \x; \btheta)$ is constructed from CNNs, and its
gradient $\gradient_\btheta E (\y, \x; \btheta)$ easily computed
by applying the chain rule as in conventional CNNs.
However, the partition function $Z$ brings difficulties for optimization.  Its gradient is written:
\begin{align}
\gradient_\btheta  \log &  Z(\x ; \btheta)  \notag  \\
	 = & \gradient_\btheta \log \sum_{\y} \exp [ -E(\y, \x) ] \notag \\
	 = &  \sum_\y \frac{\exp[-E(\y, \x; \btheta)]} {\sum_{\y'} \exp[-E(\y', \x; \btheta)]} \gradient_\btheta [ - E (\y, \x; \btheta)] \notag \\
	 = &  - \expect_{\y \sim \Prr(\y | \x ; \btheta)} \gradient_\btheta E (\y, \x; \btheta)
\end{align}
Generally the size of the output space $\cY$
is exponential in
the number of nodes, which prohibits the direct calculation of $Z$ and its gradient.
The CRF graph we considered for segmentation here is a loopy graph (not tree-structured),
in which a large number of nodes (more than $1000$) and pairwise connections (more than $ 2\times 10^4$) are involved for one image.
For loopy graph with large number of nodes and edges, typically approximation is required for inference,
and even this is generally computationally expensive.

More importantly, usually a large number of SGD iterations are required for
training CNNs. Typically the number of iterations is in tens or hundreds of thousands.
Thus performing inference at each SGD iteration is very computationally expensive.

\subsection{Piecewise training of CRFs}

Instead of directly solving the optimization in \eqref{eq:crf_learning_org},
we propose to apply an approximate CRF learning method.
In the literature,
there are two popular types of learning methods which approximate the CRF objective:
pseudo-likelihood learning \cite{besag1977efficiency} and piecewise learning \cite{SuttonM05}.
The main advantage of these methods  in term of training deep CRF is that they  do not involve marginal inference for gradient calculation,
which significantly improves the efficiency of training.
Decision tree fields \cite{NowozinRBSYK11} and regression tree fields \cite{jancsary2012regression}
are based on pseudo-likelihood learning, while
piecewise learning has been applied in the work \cite{SuttonM05,kolesnikov2014closed}.

Here we develop this idea for the case of training the CRF with the CNN potentials.
In piecewise training,
the conditional likelihood
is formulated as a number of independent likelihoods defined on potentials, written as:
\begin{align}
\Prr (\y | \x) = \prod_{U \in \cU} \prod_{ p \in \cN_{U}} \Prr_{U} ( y_p | \x) \prod_{V \in \cV}
\prod_{ (p,q) \in \cS_{V}} \Prr_{V} ( y_p, y_q | \x). \notag
\end{align}
The likelihood $\Prr_{U} ( y_p | \x)$ is constructed from the unary potential $U$.
Likewise, $\Prr_{V} ( y_p, y_q | \x)$ is constructed from the pairwise potential $V$.
$\Prr_{U}$ and $\Prr_{V}$ are written as:
\begin{align}
\Prr_{U}(y_p|\x) &  = \frac{\exp [- U(y_p, \x_p)]}{\sum_{y'_p} \exp [ -U(y'_p, \x_p) ]}, \label{eq:pu} \\
\Prr_{V}(y_p, y_q|\x) &  = \frac{\exp [- V(y_p, y_q, \x_{pq})]}{\sum_{y'_p, y'_q} \exp [ -V(y'_p, y'_q, \x_{pq}) ]}
\label{eq:pv}.
\end{align}
The log-likelihood for piecewise training is then:
\begin{align}
	\log \Prr (\y | \x)  =  &  \sum_{U \in \cU} \sum_{ p \in \cN_{U}} \log \Prr_{U} ( y_p | \x) \notag \\
	&  + \sum_{V \in \cV} \sum_{ (p,q) \in \cS_{V}} \log \Prr_{V} ( y_p, y_q | \x).
\end{align}
 The optimization problem for piecewise training is to minimize
 the negative log likelihood with regularization:
 \begin{align}
	\min_{\btheta} \frac{\lambda}{2} & \fnorm \btheta -
	 \sum_{i=1}^N \biggr[ \sum_{U \in \cU}
		\sum_{ p \in \cN_{U}^{(i)}} \log \Prr_{U} ( y_p | \x^{(i)}; \btheta_U) \notag \\
	& + \sum_{V \in \cV} \sum_{ (p,q) \in \cS_{V}^{(i)}} \log \Prr_{V} ( y_p, y_q | \x^{(i)}; \btheta_V) \biggr].
	\label{eq:partitioned}
\end{align}
Compared to the objective in \eqref{eq:crf_learning_org} for direct maximum likelihood learning,
the above objective does not involve the global partition function $Z(\x ; \btheta)$.
To calculate the gradient of the above objective,
we only need to calculate the gradient
$\gradient_{\btheta_U} \log \Prr_{U}$
and
$\gradient_{\btheta_V} \log \Prr_{V}$.
With the definition in \eqref{eq:pu}, $\Prr_{U}$ is a conventional softmax normalization function over only $K$ (the number of classes) elements.
Similar analysis can also be applied to $\Prr_{V}$.
Hence, we can easily calculate the gradient without involving expensive inference.
Moreover, we are able to perform paralleled training of potential functions, since the above objective is formulated by a summation of independent log-likelihoods.

 As previously discussed, CNN training usually involves a large number of gradient update iteration which prohibit the repeated expensive inference.
 Our piecewise approach here provides a practical solution for learning CRFs with CNN potentials on large-scale data.

\section{Implementation details}

For the FeatMap-Net,
the first $5$ convolution blocks and the first convolution layer in the $6$th convolution block are initialized from the VGG-16 network \cite{simonyan2014very}.
All remaining layers are randomly initialized.
Note that VGG-16 network is widely applied in recent segmentation methods.
All layers are trained using back-propagation/stochastic gradient descend (SGD).
We apply simple data augmentation in the training stage.
Specifically, we perform random scaling (from $0.7$ to $1.2$) and flipping of the images for training.

As illustrated in Fig.~\ref{fig:pairwise},
we use $2$ types of pairwise potential functions.
In total, we have $1$ type of unary potential function and $2$ types of pairwise potential functions.
We formulate one specific FeatMap-Net and potential network (Unary-Net or Pairwise-Net) for one type of potential function.
In other words, one type of potential function is constructed by one FeatMap-Net and a shallow potential network.
More details of FeatMap-Net and Unary/Pairwise-Net can be found in Fig.~\ref{fig:potential_net}.
There are two main benefits of modeling specific FeatMap-Net for each potential instead of sharing one FeatMap-Net across potentials.
Different types of potentials have different focus and thus probably require separate feature maps.
Using separate FeatMap-Net allows generating specific high-level features for the corresponding potential function.
Moreover, with separate FeatMap-Net, we are able to parallel the training of different types of potentials,
 and thus ease the implementation and speed up the training.

\subsection{Efficient learning}
As previously discussed, one node in the CRF graph is connected to all nodes which lie within a predefined range box.
Under this setting, the number of pairwise connections for one node can  be a few hundred.
For example, for an input image with a resolution of $500 \times 500$ pixels
and $1.2$x scaling, the resolution of the feature map after going through FeatMap-Net is around $35 \times 35$.
In this case, the number of nodes are around $1200$, and the number of connections is around $200$ for one node.
Hence for this image, we need to process $1200 \times 200$ pairwise relations for generating the edge features,
passing forward the Pairwise-Net and back-propagating the gradients (in the training stage).
These operations are considerably computationally expensive for such a large number of pairwise connections.
If using high feature dimension for the feature map, these operations can even run out of the GPU memory.

In our solution, to speed up the training and testing of the pairwise potentials,
we perform sampling of pairwise connections for each node in the CRF graph.
Since the feature map encodes redundant information in local regions, performing sampling can
still preserve sufficient pairwise relations while removing redundancies.
Specifically, we sample $24$ neighboring nodes based on a regular $5 \times 5$ grid spanning the range box (excluding self-connection),
and thus we have $24$ pairwise connections for each node, which is an order of magnitude fewer connections than the original setting.
We observe that this sampling setting, which reduces the number of pairwise connections significantly,
speeds up the training without degrading the performance.

\subsection{Asynchronous gradient update}
The number of pairwise connections is still large even with sampling, which brings the problem of keeping the edge features in the GPU memory.
Moreover, considering a large number of pairwise connections (more than $2 \times 10^4$) in one iteration for updating the parameters of Pairwise-Net
might result in degraded gradients.
This is similar to the case that using an extremely large batch size for gradient calculation in the training of a conventional classification network.
An extremely large batch size for gradient update can significantly slow down the convergence and may decrease the performance \cite{li2014efficient}.
Moreover, as discussed in \cite{bengio2012},
using small batch size may perform “noise” injection in the gradient calculation as is a form of regularization, which may lead to better parameter solutions.
Overall, from both empirical observations and theoretical analysis, a appropriate setting of batch size is key to the network training.
Therefore, we probably should not consider all pairwise connections in one gradient iteration for updating the Pairwise-Net.

To reduce the GPU memory consumption and improve the batch update for the Pairwise-Net,
we perform asynchronous gradient update for training the FeatMap-Net and  Pairwise-Net.
With asynchronous gradient update, the gradient calculations for different parts of the network are not required in the same iteration,
which breaks the dependency between different parts (or layers) of the networks.
Asynchronous gradient update is widely applied in large-scale distributed network learning.
For details one may refer to \cite{dean2012large}.

Specifically, in one stochastic gradient  iteration,
we perform multiple sub-iterations of gradient update for the Pairwise-Net and collect the gradients for the FeatMap-Net.
In each sub-iteration, a subset of pairwise connections is selected (e.g., $2000$) for gradient calculation
and the parameters of Pairwise-Net are updated.
Clearly, in each sub-iteration we only process a small number of connections for updating the network parameters of Pairwise-Net,
thus GPU consumption is low and the batch size for learning Pairwise-Net is reduced.
This asynchronous approach addresses the problems of GPU memory and large batch size for training Pairwise-Net.
After going through all pairwise connections, we collect the gradients for FeatMap-Net,
and perform a conventional back-propagation gradient update to FeatMap-Net.

%% file: exp.tex
\input{seg_example_nyud.tex}

\begin{table}[t]
\caption{Segmentation results on NYUDv2 dataset (40 classes).
We compare to a number of recent methods.
Our method significantly outperforms the existing methods.}
\centering
\resizebox{1\linewidth}{!}
  {
  \begin{tabular}{ r | c | c c c }
method	&training data	&pixel accuracy	&mean accuracy	&IoU\\ \hline \hline
Gupta et al. \cite{gupta2014learning} 	&RGB-D	&60.3	&-	&28.6\\
FCN-32s \cite{LongSD14}	&RGB	&60.0	&42.2	&29.2\\
FCN-HHA \cite{LongSD14}	&RGB-D	&65.4	&46.1	&34.0\\ \hline
ours	&RGB	&\bf 70.0	&\bf 53.6	&\bf 40.6\\
 \end{tabular}
  }
\label{tab:nyud}
\end{table}

\begin{table}[t]
\caption{Ablation Experiments. The table shows the value added
by the different system components of our method on the NYUDv2 dataset (40 classes).
}
\centering
\resizebox{1\linewidth}{!}
  {
  \begin{tabular}{ r | c c c }
method  &pixel accuracy &mean accuracy  &IoU\\ \hline \hline
FCN-32s \cite{LongSD14} &60.0 &42.2 &29.2\\ \hline
FullyConvNet Baseline  &61.5 &43.2 &30.5\\
$+$ sliding pyramid pooling  &63.5 &45.3 &32.4\\
$+$ multi-scales  &67.0 &50.1 &37.0\\
$+$ boundary refinement &68.5 &50.9 &38.3\\
$+$ CNN contextual pairwise &70.0 &53.6 &40.6\\
 \end{tabular}
  }
\label{tab:featmapnet}
\end{table}

\begin{table}[t]
\caption{Comparison with unary ensembles on the NYUDv2 dataset (40 classes). 
We compare our contextual CRF model to an ensemble of up to 4 unary-only networks.
It clearly shows that using our CRF model with 1 pairwise potential (corresponding to the surrounding relation) and 1 unary potential outperforms the ensembles of multiple unary networks.
Moreover, using an ensemble of 2 unary in our CRF model can further improve the performance (``1 pairwise + 2 unary").
These results verify the effectiveness of learning pairwise potentials.
}
\centering
\resizebox{.65\linewidth}{!}
  {
  \begin{tabular}{ r | c }
settings & IoU score\\ \hline \hline
1 unary &37.0\\
2 unary ensemble	&37.8\\
3 unary ensemble	&38.4\\
4 unary ensemble	&38.7\\ \hline
1 pairwise  + 1 unary	&38.9\\
1 pairwise  +2 unary &39.2\\
 \end{tabular}
  }
\label{tab:unary_ensemble}
\end{table}

\begin{table*}[t]
\caption{Individual category results on the PASCAL VOC 2012 test set (IoU scores). Our method performs the best}
\centering
\resizebox{1\linewidth}{!}
  {
  \begin{tabular}{ r | c c c c c c c c c c c c c c c c c c c c | c }

method & \rot{aero}  &\rot{bike} &\rot{bird} &\rot{boat} &\rot{bottle}   &\rot{bus}  &\rot{car}  &\rot{cat}  &\rot{chair}    &\rot{cow}  &\rot{table}    &\rot{dog}  &\rot{horse}    &\rot{mbike}    &\rot{person}   &\rot{potted}   &\rot{sheep}    &\rot{sofa} &\rot{train}    &\rot{tv}  & mean \\ \hline \hline
\multicolumn{22}{c}{\bf Only using VOC training data} \\ \hline
FCN-8s \cite{LongSD14}    &76.8   &34.2   &68.9   &49.4   &60.3   &75.3   &74.7   &77.6   &21.4   &62.5   &46.8   &71.8   &63.9   &76.5   &73.9   &45.2   &72.4   &37.4   &70.9 & 55.1 & 62.2 \\
Zoom-out \cite{MostajabiYS14}  &85.6 &37.3 &\bf 83.2 &62.5 &66.0 &85.1 &80.7 &84.9 &27.2 &73.2 &57.5 &78.1 &79.2 &81.1 &77.1 &53.6 &74.0 &49.2 &71.7 &63.3 &69.6\\
DeepLab \cite{ChenPKMY14} &84.4 &54.5 &81.5 &63.6 &65.9 &85.1 &79.1 &83.4 &30.7 &74.1 &59.8 &79.0 &76.1 &83.2 &80.8 &59.7 &82.2 &50.4 &73.1 &63.7 &71.6 \\
CRF-RNN \cite{zheng2015conditional} &87.5 &39.0 &79.7 &64.2 &68.3 &87.6 &80.8 &84.4 &30.4 &78.2 &60.4 &80.5 &77.8 &83.1 &80.6 &59.5 &82.8 &47.8 &78.3 &67.1 &72.0 \\
DeconvNet \cite{noh2015learning} &89.9 &39.3 &79.7 &63.9 &68.2 &87.4 &81.2 &86.1 &28.5 &77.0 &62.0 &79.0 &80.3 &83.6 &80.2 &58.8 &\bf 83.4 &54.3 &80.7 &65.0 &72.5\\
DPN \cite{LiuDPN}  &87.7  &\bf 59.4 &78.4 &64.9 &70.3 &89.3 &83.5 &86.1 &31.7 &79.9 &\bf 62.6 &81.9 &80.0 &83.5 &82.3 &60.5 &83.2 &53.4 &77.9 &65.0 &74.1\\
ours &\bf 90.6  &37.6 & 80.0 &\bf 67.8 &\bf 74.4 &\bf 92.0 &\bf 85.2 &\bf 86.2 &\bf 39.1 &\bf 81.2 & 58.9 &\bf 83.8 &\bf 83.9 &\bf 84.3 &\bf 84.8 &\bf 62.1 & 83.2 &\bf 58.2 &\bf 80.8 &\bf 72.3 & \best 75.3 \\ \hline \hline
\multicolumn{22}{c}{\bf Using VOC+COCO training data} \\ \hline
DeepLab \cite{ChenPKMY14} &89.1 &38.3 &88.1 &63.3 &69.7 &87.1 &83.1 &85.0 &29.3 &76.5 &56.5 &79.8 &77.9 &85.8 &82.4 &57.4 &84.3 &54.9 &80.5 &64.1 &72.7 \\
CRF-RNN \cite{zheng2015conditional} &90.4 &55.3 &88.7 &68.4 &69.8 &88.3 &82.4 &85.1 &32.6 &78.5 &64.4 &79.6 &81.9 &\bf 86.4 &81.8 &58.6 &82.4 &53.5 &77.4 &70.1 &74.7\\
BoxSup \cite{Dai2015arXiv} &89.8  &38.0 &\bf 89.2 &\bf 68.9 &68.0 &89.6 &83.0 &87.7 &34.4 &83.6 &\bf 67.1 &81.5 &83.7 &85.2 &83.5 &58.6 &84.9 &55.8 &\bf 81.2 &70.7 &75.2\\
DPN \cite{LiuDPN} &89.0 & \bf 61.6 &87.7 &66.8 &74.7 &91.2 &84.3 &87.6 &36.5 & 86.3 &66.1 &84.4 &87.8 &85.6 &85.4 &63.6 &87.3 &61.3 &79.4 &66.4 &77.5 \\
ours+ &\bf 94.1	&40.4	&83.6	&67.3	&\bf 75.6	&\bf 93.4	&\bf 84.4	&\bf 88.7	&\bf 41.6	&\bf 86.4	&63.3	&\bf 85.5	&\bf 89.3	&85.6	&\bf 86.0	&\bf 67.4	&\bf 90.1	&\bf 62.6	&80.9	&\bf 72.5	&\best 77.8 \\

\end{tabular}
  }
\label{tab:voc12_test_details}
\end{table*}

\input{seg_example_voc.tex}

\section{Experiments}

We evaluate our method on $8$ challenging semantic segmentation datasets:
PASCAL VOC 2012, NYUDv2, PASCAL-Context, SIFT-flow, SUN-RGBD, KITTY, COCO and Cityscapes, which covers various types of scene images,
including indoor/outdoor scene, street scene, etc.
Our comprehensive experiments show that the proposed method achieves new state-or-the-art performance on these datasets.
For VGG pre-trained layers (Block 1 to Block 5 in FeatMap-Net), we use a small learning rate: 0.0001; 
for the remaining layers (Block 6 in FeatMap-Net, layers in Unary-Net and Pairwise-Net), we set a larger learning rate: 0.001.
Our system is built on MatConvNet \cite{matconvnet}.

The segmentation performance is measured by the intersection-over-union (IoU) score \cite{everingham2010pascal},
the pixel accuracy and the mean accuracy on categories \cite{LongSD14}.
We denote $c_{ij}$ as an element in the confusion matrix,
which is the number of pixels with the $i$-th category as the ground truth and
the $j$-th category as the prediction; $t_i$ is the total number of pixels for the $i$-th category in the ground truth;
$K$ is the number of categories.
Pixel accuracy measures the portion of correctly predicted pixels:
$\frac{\sum_{i} c_{ii}}{\sum_i t_i} $.
Mean accuracy measures the per-category pixel accuracy: $ \frac{1}{K} \sum_{i} \frac{c_{ii}}{t_i} $.
IoU score calculates the portion of the intersection
between the ground truth and the prediction: $ \frac{1}{K} \sum_{i} \frac{c_{ii}}{t_i + \sum_j c_{ij} - c_{ii}} $.

\subsection{Results on the NYUDv2 dataset}

We first evaluate our method on the  NYUDv2 \cite{silberman2012indoor} dataset which
has $1449$ RGB-D indoor scene images.
We use the segmentation labels provided in \cite{gupta2013perceptual} for which the labels are processed into $40$ classes.
We use the standard training set which contains $795$ images and the test set which contains $654$ images.
We train our models only on RGB images without using the depth information.

Results are shown in Table \ref{tab:nyud}.
Some prediction examples are shown in Fig.~\ref{fig:example_nyud}
Unless otherwise specified, our models are initialized using the VGG-16 network.
VGG-16 is also used in the competing method FCN \cite{LongSD14}.
our contextual model with CNN pairwise potentials achieves the best performance, which sets new state-of-the-art result on the NYUDv2 dataset, 
Note that we do not use any depth information in our model.

\subsubsection{Component evaluation}
\label{sec:nyud_com}

We evaluate the performance contribution of different components of the FeatMap-Net for capture patch-background context on the NYUDv2 dataset.
We present the results of adding different components in FeatMap-Net, which are shown in Table \ref{tab:featmapnet}.
We start from a baseline setting of our FeatMap-Net (``FullyConvNet Baseline" in the result table), for which multi-scale and sliding pooling is removed.
This baseline setting is the conventional fully convolution network
for segmentation, which can be considered as our implementation of the FCN method in \cite{LongSD14}.
The result shows that our CNN baseline implementation (``FullyConvNet") achieves very similar performance (slightly better) than the FCN method.
Applying multi-scale network design and sliding pyramid pooling significantly improve the performance,
which clearly shows the benefits of encoding rich background context in our approach.
Applying the dense CRF method \cite{krahenbuhl2012efficient} for boundary refinement gains further improvement.
Finally, adding our contextual CNN pairwise potentials brings significant further improvement, for which we achieve the best performance in this dataset.

\subsubsection{Comparison with multi-unary ensemble}
\label{sec:nyud_ensemble}

We compare our CRF model with contextual pairwise potentials against the simple ensemble of multiple
unary-only models. Four unary-only networks are independently trained in this experiment. Results are shown in Table \ref{tab:unary_ensemble}.
To clearly evaluate the effectiveness, 
we use 1 type of pairwise potential which corresponds to the surrounding relations in our CRF model.
The result shows that using our CRF model with 1 pairwise potential and 1 unary potential (``1 pairwise + 1 unary") outperforms the ensembles of multiple unary networks,
which verifies the effectiveness of learning pairwise potentials.
Moreover, using extra unary networks, i.e., an ensemble of 2 unary networks, in our CRF model can further improve the performance, as shown by the entry ``1 pairwise + 2 unary" in the result table.
It indicates that our pairwise potential is able to capture different information and complementary to the multi-unary ensemble.

\subsection{Results on the PASCAL VOC 2012 dataset}
\label{sec:voc2012exp}

PASCAL VOC 2012 \cite{everingham2010pascal} is a well-known segmentation evaluation dataset which consists of 20 object categories and one background category.
This dataset is split into a training set, a validation set and a test set,
which respectively contain $1464$, $1449$ and $1456$ images.
Following a conventional setting in \cite{BharathECCV2014,ChenPKMY14}, the training set is augmented by extra annotated VOC images provided in \cite{HariharanABMM11}, which results in $10582$ training images.
We verify our performance on the PASCAL VOC 2012 test set.
We compare with a number of recent methods with competitive  performance.
Since the ground truth labels are not available for the test set,
we evaluate our method through the VOC evaluation server.

The IoU scores are shown in the last column of Table \ref{tab:voc12_test_details}.
Prediction examples of our method are shown in Fig. \ref{fig:example_voc}.
We first train our model only using the VOC images.
We achieve an IoU score of $75.3$, which is the best result amongst methods that only use the VOC training data.\footnote{The result link at the VOC evaluation server:
\url{http://host.robots.ox.ac.uk:8080/anonymous/KEFFM4.html}}

To improve the performance, following the setting in recent work \cite{ChenPKMY14,Dai2015arXiv},
we train our model with the extra images from the COCO dataset \cite{lin2014microsoft}.
With these extra training images, we achieve an IoU score of $77.2$.

As described in Sec.~\ref{sec:prediction},
our deep structured model  generates low-resolution coarse prediction, which is $1/16$ of the input image size.
To obtain the final high-resolution prediction we apply a simple yet effective approach:
we first perform bilinear upsampling of the coarse score map and then apply the boundary refinement post-processing \cite{krahenbuhl2012efficient}.
To improving this simple approach,
we exploit the feature maps from middle layers to refine the coarse prediction
and produce high-resolution prediction, which is similar to
the methods in \cite{ChenPKMY14,LongSD14,hariharan2014hypercolumns}.
With this improved refinement approach,
we finally achieve an IoU score of $77.8$, {\em which is best reported result on this challenging dataset.}
\footnote{The result link at the VOC evaluation server: \url{http://host.robots.ox.ac.uk:8080/anonymous/MVTNTX.html}}

\begin{figure}[t]
	\center
	\includegraphics[width=1\linewidth]{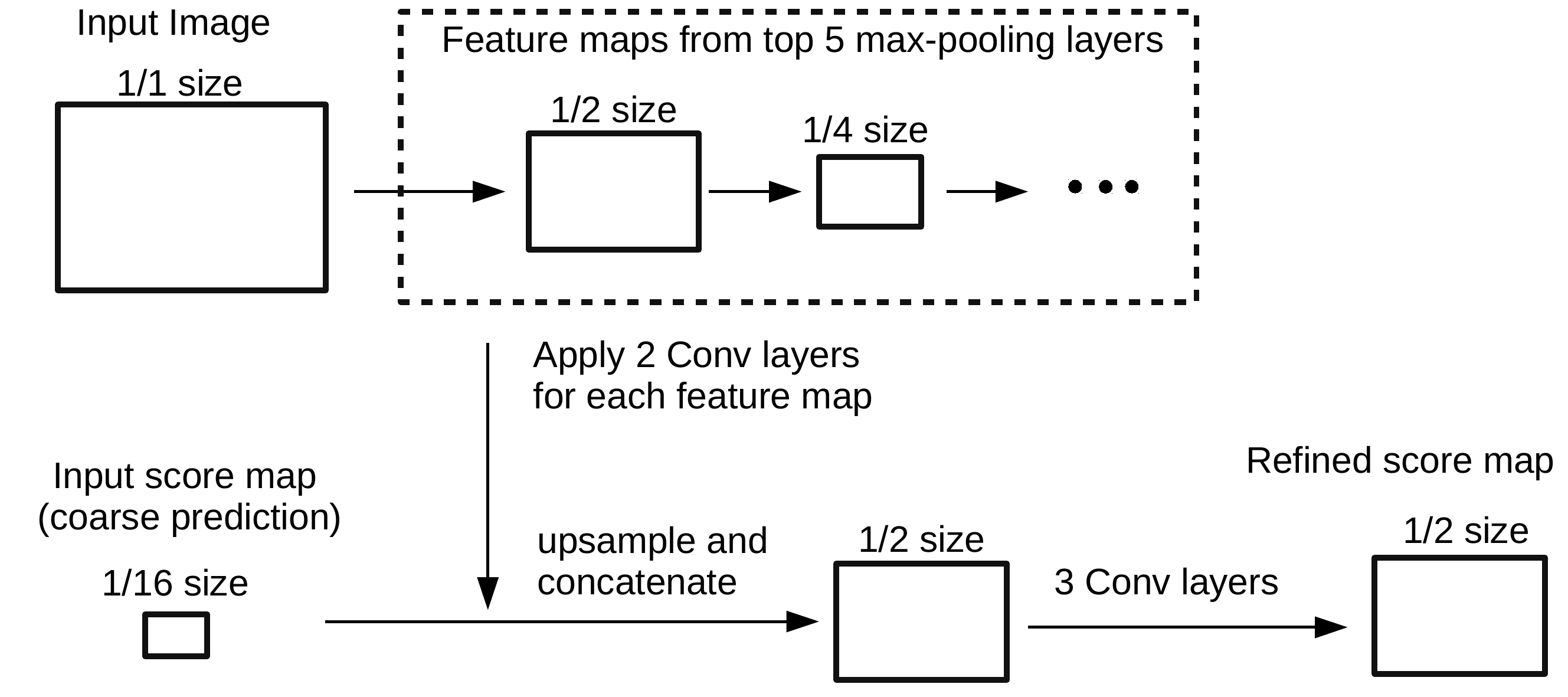}
\caption{
The illustration of exploiting the feature maps from middle layers to refine the low-resolution ($1/16$ of the input image) coarse prediction.
The refined prediction has a resolution of $1/2$ of the input image.
}
\label{fig:midlayer}
\end{figure}

The feature maps from the middle layers encode lower level visual information (from edge patterns to texture/object part patterns)
and have higher resolution than the final output, thus it is  expected that learning extra layers on these feature maps
helps to predict details in the object boundaries.
Specifically, we add refinement layers
on top of the feature maps from the first $5$ max-pooling layers and the score map of the coarse prediction (output by our deep structured model).
Details are shown in Fig.~\ref{fig:midlayer}.
These refinement layers play a role of refining the coarse prediction by exploring middle layer features,
which increase the resolution of the prediction from $1/16$ (coarse prediction) to $1/2$ of the input image.
With this improved prediction, we perform boundary refinement using \cite{krahenbuhl2012efficient} to generate the final prediction.

The results for each category are shown in Table \ref{tab:voc12_test_details}.
We outperform comparing methods in most categories.
For only using the VOC training set, our method outperforms the second best method, DPN \cite{LiuDPN},
on $18$ categories out of $20$.
For using VOC+COCO training set, our method outperforms DPN \cite{LiuDPN}
on $15$ categories out of $20$.

\input{seg_example_city.tex}

\begin{table}[t]
\caption{Segmentation results on the Cityscapes \emph{test} set.
our method achieves the best performance.
}
\centering
\resizebox{.5\linewidth}{!}
  {
  \begin{tabular}{ r |  c }
Method	& IoU score \\ \hline \hline
FCN-8s \cite{LongSD14} & 65.3  \\
DPN \cite{LiuDPN} & 66.8 \\
Dilation10 \cite{YuK15}  & 67.1 \\
DeepLab-CRF \cite{ChenPKMY14} & 63.1 \\
ours &  \bf 71.6
 \end{tabular}
  }
\label{tab:cityscape}
\end{table}

\subsection{Results on the Cityscapes dataset}
\label{sec:city}

The large scale outdoor image dataset Cityscapes \cite{Cordts2016Cityscapes}
contains high-resolution street scene images from $50$ different cities.
This dataset provides pixel-level semantic segmentation labels of $5000$ images for $25$ classes including road, car, pedestrian, bicycle, sky etc.
The provided ``trainval" set has $3475$ image.
We use the training set ($2975$ images) for training.
The ground truth of the test set is not available,
and we evaluate our method through their evaluation server.
We follow the provided protocol for dataset evaluation: $19$ classes are valid for evaluation, and the remaining $6$ classes are not considered in evaluation.

Results are shown in Table \ref{tab:cityscape}. 
As similar to the setting for the PASCAL VOC dataset, we train a refinement network which is described in Fig.~\ref{fig:midlayer} to obtain high resolution prediction.
Here we set the output resolution in the refinement network as $1/4$ of the input image size.
The result clearly shows that our method outperforms other competing methods.
Prediction examples on the validation set are shown in Fig.~\ref{fig:example_city}.

\begin{table}[t]
\caption{Segmentation results on PASCAL-Context dataset (60 classes).
Our method performs the best.}
\centering
\resizebox{.8\linewidth}{!}
  {
  \begin{tabular}{ r | c c c}
  method	&pixel accuracy	&mean accuracy	&IoU\\ \hline \hline
O2P \cite{carreira2012semantic}	&-	&-	&18.1\\
CFM \cite{dai2014convolutional}	&-	&-	&34.4\\
FCN-8s \cite{LongSD14}	&65.9	&46.5	&35.1\\
BoxSup \cite{Dai2015arXiv}	&-	&-	&40.5\\ \hline
ours	&\bf 71.5	&\bf 53.9	&\bf 43.3\\
 \end{tabular}
  }
\label{tab:pascalcontext}
\end{table}

\subsection{Results on the PASCAL-Context dataset}
The
PASCAL-Context \cite{mottaghi2014role} dataset provides the segmentation labels of the whole scene (including the ``stuff" labels) for the PASCAL VOC images.
We use the segmentation labels which contain $60$ classes ($59$ classes plus the `` background" class ) for evaluation.
We use the provided training/test splits.
The training set contains $4998$ images and the test set contains $5105$ images.

Results are shown in Table \ref{tab:pascalcontext}.
Prediction examples are shown in Fig. \ref{fig:example_pascalcontext}
Our method significantly outperform the competing methods. To our knowledge, {\em ours is the best reported result on this dataset}.

\subsection{Results on the SUN-RGBD dataset}
SUN-RGBD \cite{song2015sun} is a segmentation dataset contains around $10,000$ indoor images and provides pixel labeling masks of $37$ classes,
which is an extension of the NYUD dataset \cite{silberman2012indoor}.
Results are shown in Table \ref{tab:sunrgbd}. Our method outperform
the existing methods by a large margin, even though we does not make use of the depth information for training.

\begin{table}[t]
\caption{Segmentation results on SUN-RGBD dataset (37 classes).
We compare to a number of recent methods.
Our method significantly outperforms the existing methods.}
\centering
\resizebox{1\linewidth}{!}
  {
  \begin{tabular}{ r | c | c c c }
  method  &training data  &pixel accuracy &mean accuracy  &IoU\\ \hline \hline
  Liu et al. \cite{liu2011sift}  &RGB-D  & $-$ &10.0 & $-$ \\
  Ren et al. \cite{ren2012rgb}  &RGB-D  & $-$ &36.3 & $-$ \\
  Kendall et al. \cite{KendallBC15} &RGB  &71.2 &45.9 &30.7 \\ \hline
  ours  &RGB  &\bf 78.4 &\bf 53.4 &\bf 42.3 \\
 \end{tabular}
  }
\label{tab:sunrgbd}
\end{table}

\subsection{Results on the COCO dataset}

The COCO dataset \cite{lin2014microsoft} contains more than 1 million images and provide segmentation labels for $80$ classes.
Since the test set is not available, we generate $2599$ images for testing and the remaining images are for training.
We select the test images on a class balance basis which ensures every category at lease appears in $50$ images.
Labeling regions which are smaller than $200$ pixels are treated as ``void" which are not considered in training and evaluation.
Results are shown in Table \ref{tab:cocofull}.
We compared to two baseline methods which are based on conventional fully convolution networks.
The details of these baseline methods are the same as that for the Cityscapes dataset (see Sec. \ref{sec:city}).
The results shows that our method significantly outperforms the baselines.

\begin{table}[t]
\caption{Segmentation results on COCO dataset (80 classes).
Our method significantly outperforms the fully convolution network (``FullyConvNet"). }
\centering
\resizebox{1\linewidth}{!}
  {
  \begin{tabular}{ r | c c c }
  method   &pixel accuracy &mean accuracy  &IoU\\ \hline \hline
  FullyConvNet  &84.2 &56.9 &37.2\\
FullyConvNet + refine &86.7 &55.0 &41.3\\
ours  &\bf 88.3 &\bf 58.7 &\bf 46.8\\
 \end{tabular}
  }
\label{tab:cocofull}
\end{table}

\begin{table}[t]
\caption{Segmentation results on SIFT-flow dataset (33 classes).
Our method performs the best.}
\centering
\resizebox{1\linewidth}{!}
  {
  \begin{tabular}{ r | c c c}
  method	&pixel accuracy	&mean accuracy	&IoU\\ \hline \hline
Liu et al. \cite{liu2011sift}	&76.7	&-	&-\\
Tighe et al. \cite{tighe2013finding} 	&75.6	&41.1	&-\\
Tighe et al. (MRF) \cite{tighe2013finding}	&78.6	&39.2	&-\\
Farabet et al. (balance) \cite{farabet2013learning}	&72.3	&50.8	&-\\
Farabet et al. \cite{farabet2013learning} 	&78.5	&29.6	&-\\
Pinheiro et al. \cite{pinheiro2013recurrent} &77.7 &29.8 &-\\
FCN-16s \cite{LongSD14}	&85.2	&51.7	&39.5\\ \hline
ours	&\bf 88.1	&\bf 53.4	&\bf 44.9\\
 \end{tabular}
  }
\label{tab:siftflow}
\end{table}

\subsection{Results on the SIFT-flow dataset}

We further evaluate our method on the SIFT-flow dataset.
This dataset contains $2688$ images and provides the segmentation labels for $33$ classes.
We use the standard split for training and evaluation.
The training set has $2488$ images and the test set has $200$ images.
Since the images are in small sizes, we upscale the image by a factor of $2$ for training.
Results are shown in Table \ref{tab:siftflow}. We achieve the best performance on this dataset.

\subsection{Results on the KITTI dataset}
We perform further evaluation on the KITTI dataset \cite{Geiger2012CVPR} for road image segmentation.
Zhang at el. \cite{zhang2015ICRA} provide semantic segmentation labels of $10$ classes for $252$ images, in which $140$ images are for training and the remaining $112$ are for testing.
We follow the provided training and testing splits for evaluation and report the results in Table \ref{tab:kitti}.
Clearly, our method performs the best.
To further improve the performance, we perform pre-training on COCO images,
for which the result is denoted by ``ours+" in the result table.

\begin{table}[t]
\caption{Segmentation results on KITTI dataset (10 classes).
We compare to a number of recent methods.
Our method significantly outperforms the existing methods.}
\centering
\resizebox{1\linewidth}{!}
  {
  \begin{tabular}{ r | c c c }
  method   &pixel accuracy &mean accuracy  &IoU\\ \hline \hline
  Cadena et al. \cite{cadena2014semantic}   &84.1 &52.4 & $-$\\
Zhang et al. \cite{zhang2015ICRA}  &89.3 &65.4 & $-$ \\ \hline
ours  &93.3 &74.5 &68.5\\
ours+ &\bf 94.3 &\bf 75.9 &\bf 70.3\\
 \end{tabular}
  }
\label{tab:kitti}
\end{table}

\input{seg_example_pascalcontext_single.tex}

%% file: seg_example_nyud.tex
\ifx\imglistflag\undefined
\def\imglistflag{}
\newcounter{imgidx}
\newcounter{cntone}
\newcounter{cnttwo}
\newcounter{img_total_one}
\newcounter{img_total_two}
\newcounter{cntthree}
\newcounter{img_total_three}
\fi

 \setcounter{cntone}{0}
 \setcounter{cnttwo}{0}
 \setcounter{img_total_one}{0}
 \setcounter{img_total_two}{0}

\providecommand\settextone[2]{%
  \csdef{textone#1}{#2}}
\providecommand\addtextone[1]{%
  \stepcounter{cntone}%
  \csdef{textone\thecntone}{#1}}
\providecommand\gettextone[1]{%
  \csuse{textone#1}}

\providecommand\settexttwo[2]{%
  \csdef{texttwo#1}{#2}}
\providecommand\addtexttwo[1]{%
  \stepcounter{cnttwo}%
  \csdef{texttwo\thecnttwo}{#1}}
\providecommand\gettexttwo[1]{%
  \csuse{texttwo#1}}

\addtextone{000413}
\addtextone{001096}
\addtextone{001258}
\addtextone{001299}
\addtextone{001431}

\setcounter{img_total_one}{\arabic{cntone}}
\stepcounter{img_total_one}

\setcounter{img_total_two}{\arabic{cnttwo}}
\stepcounter{img_total_two}

\begin{figure}[t]
\centering
 \resizebox{.975\linewidth}{!} {
    \begin{subfigure}{1.2in}
	\forloop{imgidx}{1}{\value{imgidx} < \value{img_total_one}}{
		\includegraphics[width=1.2in]{examples/nyud_extra/{\gettextone{\arabic{imgidx}}}.png}\vspace{2pt}
	}
    \caption{Testing}
    \end{subfigure}\hspace{1pt}
    \centering
    \begin{subfigure}{1.2in}
	\forloop{imgidx}{1}{\value{imgidx} < \value{img_total_one}}{
		\includegraphics[width=1.2in]{examples/nyud_extra/{gt_\gettextone{\arabic{imgidx}}}.png}\vspace{2pt}
	}
    \caption{Ground Truth}
    \end{subfigure}\hspace{1pt}
    \centering
    \begin{subfigure}{1.2in}
    \forloop{imgidx}{1}{\value{imgidx} < \value{img_total_one}}{
		\includegraphics[width=1.2in]{examples/nyud_extra/{pred_\gettextone{\arabic{imgidx}}}.png}\vspace{2pt}
	}
    \caption{Prediction}
    \end{subfigure}
 }
    \caption{Prediction examples on the NYUDv2 dataset.}
    \label{fig:example_nyud}
\end{figure}

%% file: seg_example_voc.tex
\ifx\imglistflag\undefined
\def\imglistflag{}
\newcounter{imgidx}
\newcounter{cntone}
\newcounter{cnttwo}
\newcounter{cntthree}
\newcounter{img_total_one}
\newcounter{img_total_two}
\newcounter{img_total_three}
\fi

\setcounter{cntone}{0}
\setcounter{cnttwo}{0}
\setcounter{cntthree}{0}
\setcounter{img_total_one}{0}
\setcounter{img_total_two}{0}
\setcounter{img_total_three}{0}

\providecommand\settextone[2]{%
  \csdef{textone#1}{#2}}
\providecommand\addtextone[1]{%
  \stepcounter{cntone}%
  \csdef{textone\thecntone}{#1}}
\providecommand\gettextone[1]{%
  \csuse{textone#1}}

\providecommand\settexttwo[2]{%
  \csdef{texttwo#1}{#2}}
\providecommand\addtexttwo[1]{%
  \stepcounter{cnttwo}%
  \csdef{texttwo\thecnttwo}{#1}}
\providecommand\gettexttwo[1]{%
  \csuse{texttwo#1}}

\providecommand\settextthree[2]{%
  \csdef{textthree#1}{#2}}
\providecommand\addtextthree[1]{%
  \stepcounter{cntthree}%
  \csdef{textthree\thecntthree}{#1}}
\providecommand\gettextthree[1]{%
  \csuse{textthree#1}}

\addtextone{2007_001311}
\addtextone{2007_001284}
\addtextone{2007_001430}
\addtextone{2008_000149}
\addtextone{2007_007470}
\addtextone{2007_000762}
\addtextone{2008_000533}

\addtexttwo{2010_000666}
\addtexttwo{2007_000830}
\addtexttwo{2007_009346}
\addtexttwo{2009_003666}
\addtexttwo{2007_002624}
\addtexttwo{2010_005860}
\addtexttwo{2008_003333}

\addtextthree{2009_003703}
\addtextthree{2008_006752}
\addtextthree{2010_000372}
\addtextthree{2009_003071}
\addtextthree{2007_000346}
\addtextthree{2011_000455}
\addtextthree{2008_000661}

\setcounter{img_total_one}{\arabic{cntone}}
\stepcounter{img_total_one}

\setcounter{img_total_two}{\arabic{cnttwo}}
\stepcounter{img_total_two}

\setcounter{img_total_three}{\arabic{cntthree}}
\stepcounter{img_total_three}

\begin{figure*}[t]
\centering
\resizebox{.9985\linewidth}{!} {
    \begin{subfigure}{1.0in}
    \centering
    \forloop{imgidx}{1}{\value{imgidx} < \value{img_total_one}}{
        \includegraphics[width=1.0in,height=.8in,keepaspectratio]
        {examples/voc2012_val/{\gettextone{\arabic{imgidx}}}.png}\vspace{2pt}
    }
    \vskip -7pt
    \caption{Testing}
    \end{subfigure}\hspace{1pt}
    \begin{subfigure}{1.0in}
    \centering
    \forloop{imgidx}{1}{\value{imgidx} < \value{img_total_one}}{
        \includegraphics[width=1.0in,height=.8in,keepaspectratio]
        {examples/voc2012_val/{gt_\gettextone{\arabic{imgidx}}}.png}\vspace{2pt}
    }
    \vskip -7pt
    \caption{Ground Truth}
    \end{subfigure}\hspace{1pt}
    \begin{subfigure}{1.0in}
    \centering
    \forloop{imgidx}{1}{\value{imgidx} < \value{img_total_one}}{
        \includegraphics[width=1.0in,height=.8in,keepaspectratio]
        {examples/voc2012_val/{pred_\gettextone{\arabic{imgidx}}}.png}\vspace{2pt}
    }
    \vskip -7pt
    \caption{Prediction}
    \end{subfigure}\hspace{1pt}

    \hfill\vrule\hfill\hspace{1pt}

    \begin{subfigure}{1.0in}
    \centering
    \forloop{imgidx}{1}{\value{imgidx} < \value{img_total_two}}{
        \includegraphics[width=1.0in,height=.8in,keepaspectratio]
        {examples/voc2012_val/{\gettexttwo{\arabic{imgidx}}}.png}\vspace{2pt}
    }
    \vskip -7pt
    \caption{Testing}
    \end{subfigure}\hspace{1pt}
    \begin{subfigure}{1.0in}
    \centering
    \forloop{imgidx}{1}{\value{imgidx} < \value{img_total_two}}{
        \includegraphics[width=1.0in,height=.8in,keepaspectratio]
        {examples/voc2012_val/{gt_\gettexttwo{\arabic{imgidx}}}.png}\vspace{2pt}
     }
     \vskip -7pt
     \caption{Ground Truth}
     \end{subfigure}\hspace{1pt}
     \begin{subfigure}{1.0in}
     \centering
     \forloop{imgidx}{1}{\value{imgidx} < \value{img_total_two}}{
        \includegraphics[width=1.0in,height=.8in,keepaspectratio]
        {examples/voc2012_val/{pred_\gettexttwo{\arabic{imgidx}}}.png}\vspace{2pt}
     }
     \vskip -7pt
     \caption{Prediction}
     \end{subfigure} \hspace{1pt}

      \hfill\vrule\hfill\hspace{1pt}

    \begin{subfigure}{1.0in}
    \centering
    \forloop{imgidx}{1}{\value{imgidx} < \value{img_total_three}}{
        \includegraphics[width=1.0in,height=.8in,keepaspectratio]
        {examples/voc2012_val/{\gettextthree{\arabic{imgidx}}}.png}\vspace{2pt}
    }
    \vskip -7pt
    \caption{Testing}
    \end{subfigure}\hspace{1pt}
    \begin{subfigure}{1.0in}
    \centering
    \forloop{imgidx}{1}{\value{imgidx} < \value{img_total_three}}{
        \includegraphics[width=1.0in,height=.8in,keepaspectratio]
        {examples/voc2012_val/{gt_\gettextthree{\arabic{imgidx}}}.png}\vspace{2pt}
     }
     \vskip -7pt
     \caption{Ground Truth}
     \end{subfigure}\hspace{1pt}
     \begin{subfigure}{1.0in}
     \centering
     \forloop{imgidx}{1}{\value{imgidx} < \value{img_total_three}}{
        \includegraphics[width=1.0in,height=.8in,keepaspectratio]
        {examples/voc2012_val/{pred_\gettextthree{\arabic{imgidx}}}.png}\vspace{2pt}
     }
     \vskip -7pt
     \caption{Predict}
     \end{subfigure}

}
\vspace{2pt}
    \caption{Some prediction examples of our method on the PASCAL VOC 2012 dataset.}
    \label{fig:example_voc}
\end{figure*}

%% file: seg_example_city.tex
\ifx\imglistflag\undefined
\def\imglistflag{}
\newcounter{imgidx}
\newcounter{cntone}
\newcounter{cnttwo}
\newcounter{img_total_one}
\newcounter{img_total_two}
\fi

 \setcounter{cntone}{0}
 \setcounter{cnttwo}{0}
 \setcounter{img_total_one}{0}
 \setcounter{img_total_two}{0}

\providecommand\settextone[2]{%
  \csdef{textone#1}{#2}}
\providecommand\addtextone[1]{%
  \stepcounter{cntone}%
  \csdef{textone\thecntone}{#1}}
\providecommand\gettextone[1]{%
  \csuse{textone#1}}

\providecommand\settexttwo[2]{%
  \csdef{texttwo#1}{#2}}
\providecommand\addtexttwo[1]{%
  \stepcounter{cnttwo}%
  \csdef{texttwo\thecnttwo}{#1}}
\providecommand\gettexttwo[1]{%
  \csuse{texttwo#1}}

\addtextone{frankfurt00001_005410}
\addtextone{frankfurt00001_032018}
\addtextone{frankfurt00001_041517}
\addtextone{frankfurt00001_046779}
\addtextone{frankfurt00001_055172}

\addtexttwo{munster00009_000019}
\addtexttwo{munster00033_000019}
\addtexttwo{munster00036_000019}
\addtexttwo{munster00044_000019}
\addtexttwo{munster00056_000019}

\setcounter{img_total_one}{\arabic{cntone}}
\stepcounter{img_total_one}

\setcounter{img_total_two}{\arabic{cnttwo}}
\stepcounter{img_total_two}

\begin{figure*}[t]
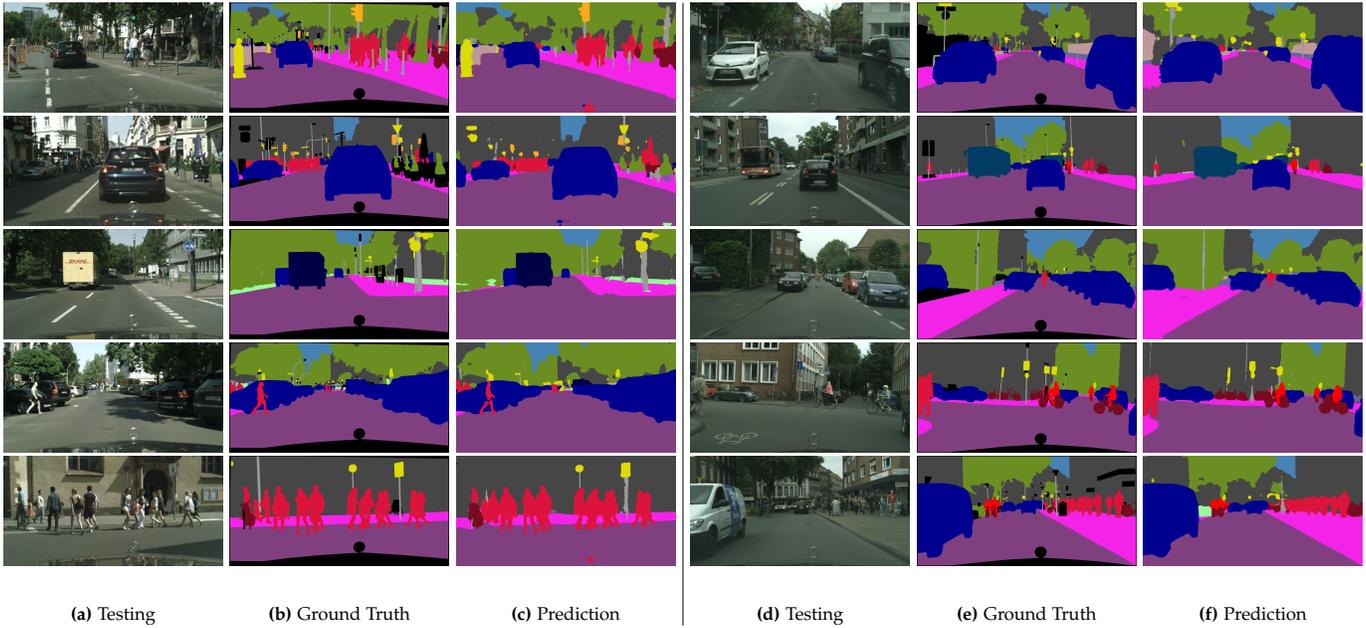

\centering
\resizebox{1\linewidth}{!} {
    \begin{subfigure}{1.5in}
	\forloop{imgidx}{1}{\value{imgidx} < \value{img_total_one}}{
		\includegraphics[width=1.5in]{examples/city/{\gettextone{\arabic{imgidx}}}.png}\vspace{2pt}
	}
    \caption{Testing}
    \end{subfigure}\hspace{1pt}
    \centering
    \begin{subfigure}{1.5in}
	\forloop{imgidx}{1}{\value{imgidx} < \value{img_total_one}}{
		\includegraphics[width=1.5in]{examples/city/{gt_\gettextone{\arabic{imgidx}}}.png}\vspace{2pt}
	}
    \caption{Ground Truth}
    \end{subfigure}\hspace{1pt}
    \centering
    \begin{subfigure}{1.5in}
    \forloop{imgidx}{1}{\value{imgidx} < \value{img_total_one}}{
		\includegraphics[width=1.5in]{examples/city/{pred_\gettextone{\arabic{imgidx}}}.png}\vspace{2pt}
	}
    \caption{Prediction}
    \end{subfigure}\hspace{1pt}

     \hfill\vrule\hfill\hspace{1pt}

    \begin{subfigure}{1.5in}
 	\forloop{imgidx}{1}{\value{imgidx} < \value{img_total_two}}{
		\includegraphics[width=1.5in]{examples/city/{\gettexttwo{\arabic{imgidx}}}.png}\vspace{2pt}
	}
    \caption{Testing}
    \end{subfigure}\hspace{1pt}
    \centering
    \begin{subfigure}{1.5in}
  	\forloop{imgidx}{1}{\value{imgidx} < \value{img_total_two}}{
	 	\includegraphics[width=1.5in]{examples/city/{gt_\gettexttwo{\arabic{imgidx}}}.png}\vspace{2pt}
	 }
     \caption{Ground Truth}
     \end{subfigure}\hspace{1pt}
     \centering
     \begin{subfigure}{1.5in}
     \forloop{imgidx}{1}{\value{imgidx} < \value{img_total_two}}{
	 	\includegraphics[width=1.5in]{examples/city/{pred_\gettexttwo{\arabic{imgidx}}}.png}\vspace{2pt}
	 }
     \caption{Prediction}
     \end{subfigure}
}
    \caption{Prediction examples of our method on Cityscapes dataset.}
    \label{fig:example_city}
\end{figure*}

%% file: seg_example_pascalcontext_single.tex
\ifx\imglistflag\undefined
\def\imglistflag{}
\newcounter{imgidx}
\newcounter{cntone}
\newcounter{cnttwo}
\newcounter{img_total_one}
\newcounter{img_total_two}
\fi

 \setcounter{cntone}{0}
 \setcounter{cnttwo}{0}
 \setcounter{img_total_one}{0}
 \setcounter{img_total_two}{0}

\providecommand\settextone[2]{%
  \csdef{textone#1}{#2}}
\providecommand\addtextone[1]{%
  \stepcounter{cntone}%
  \csdef{textone\thecntone}{#1}}
\providecommand\gettextone[1]{%
  \csuse{textone#1}}

\providecommand\settexttwo[2]{%
  \csdef{texttwo#1}{#2}}
\providecommand\addtexttwo[1]{%
  \stepcounter{cnttwo}%
  \csdef{texttwo\thecnttwo}{#1}}
\providecommand\gettexttwo[1]{%
  \csuse{texttwo#1}}

\addtextone{2008_001231}
\addtextone{2008_004887}
\addtextone{2010_000754}
\addtextone{2009_001196}
\addtextone{2010_000524}

\setcounter{img_total_one}{\arabic{cntone}}
\stepcounter{img_total_one}

\setcounter{img_total_two}{\arabic{cnttwo}}
\stepcounter{img_total_two}

\begin{figure*}[t]
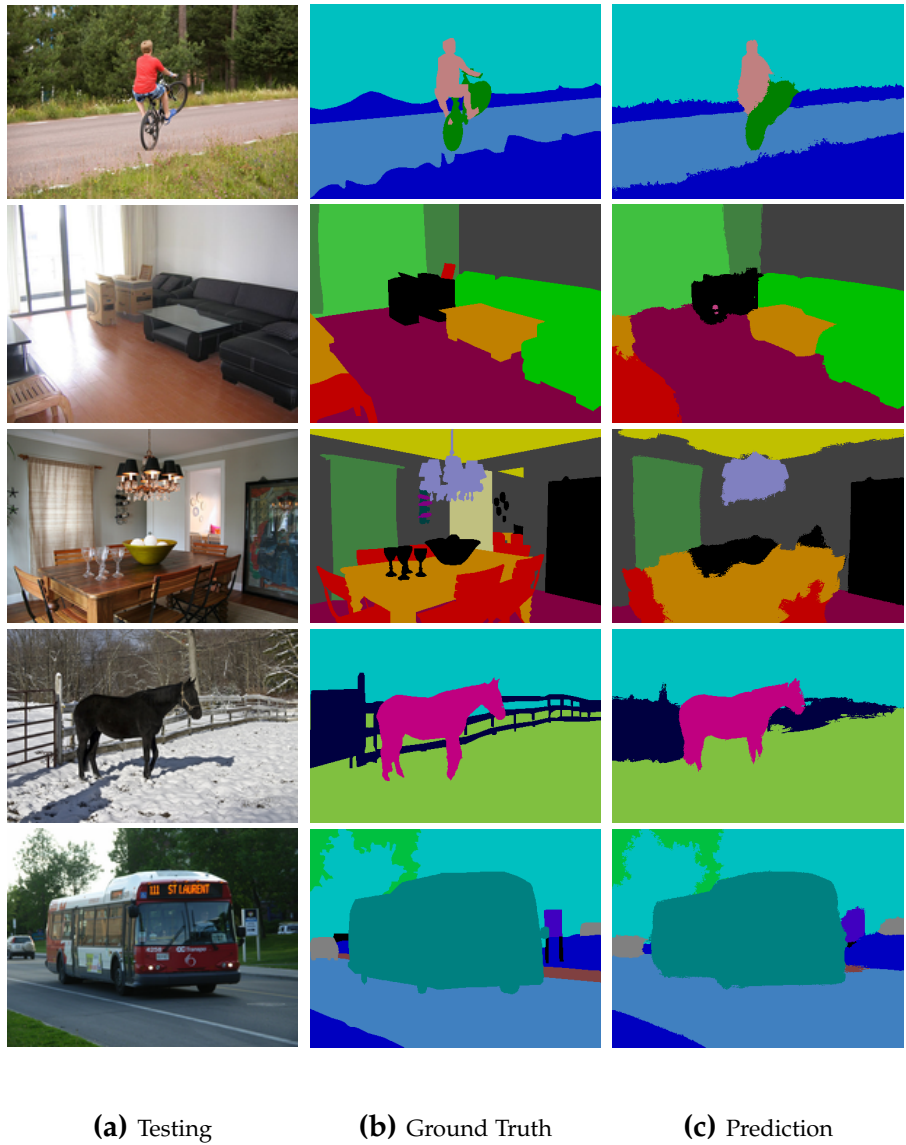

\centering
\resizebox{.67\linewidth}{!} {
    \begin{subfigure}{1.2in}
	\forloop{imgidx}{1}{\value{imgidx} < \value{img_total_one}}{
		\includegraphics[width=1.2in]{examples/pascalcontext_extra/{context_\gettextone{\arabic{imgidx}}}.png}\vspace{2pt}
	}
    \caption{\scriptsize Testing}
    \end{subfigure}\hspace{1pt}
    \centering
    \begin{subfigure}{1.2in}
	\forloop{imgidx}{1}{\value{imgidx} < \value{img_total_one}}{
		\includegraphics[width=1.2in]{examples/pascalcontext_extra/{context_gt_\gettextone{\arabic{imgidx}}}.png}\vspace{2pt}
	}
    \caption{\scriptsize Ground Truth}
    \end{subfigure}\hspace{1pt}
    \centering
    \begin{subfigure}{1.2in}
    \forloop{imgidx}{1}{\value{imgidx} < \value{img_total_one}}{
		\includegraphics[width=1.2in]{examples/pascalcontext_extra/{context_pred_\gettextone{\arabic{imgidx}}}.png}\vspace{2pt}
	}
    \caption{\scriptsize Prediction}
    \end{subfigure}\hspace{1pt}
}
    \caption{Prediction examples on the PASCAL-Context dataset.}
    \label{fig:example_pascalcontext}
\end{figure*}